\documentclass[authoryear,preprint,review,12pt]{elsarticle}

\usepackage{amssymb}
\usepackage{amsmath}

\usepackage{booktabs}
\usepackage{multirow}
\usepackage{hyperref}
\usepackage{xcolor}
\usepackage{subfig}
\usepackage{soul}

\newcommand{\vecx}{\mathbf{x}}
\newcommand{\vecg}{\mathbf{g}}
\newcommand{\vecy}{\mathbf{y}}
\newcommand{\veczero}{\mathbf{0}}
\newcommand{\vecone}{\mathbf{1}}

\newcommand{\vecz}{\mathbf{z}}

\newcommand{\vectheta}{\boldsymbol{\theta}}
\newcommand{\vecomega}{\boldsymbol{\omega}}

\journal{JAG}

\begin{document}

\begin{frontmatter}



\title{Generative Adversarial Models for Extreme Geospatial Downscaling} 


\author{Guiye Li} 
\ead{Guiye.Li@colorado.edu}

\author{Guofeng Cao\corref{cor1}}
\cortext[cor1]{Corresponding author}
\ead{Guofeng.Cao@colorado.edu}

\address{Department of Geography, University of Colorado Boulder, Boulder, CO, 80309, USA}

\begin{abstract}
Addressing the challenges of climate change requires accurate and
high-resolution mapping of geospatial data, especially climate and weather
variables. However, many existing geospatial datasets, such as the gridded
outputs of the state-of-the-art numerical climate models (e.g., general
circulation models), are only available at very coarse spatial resolutions
due to the model complexity and extremely high computational demand.
Deep-learning-based methods, particularly generative adversarial networks
(GANs) and their variants, have proved effective for refining natural
images and have shown great promise in improving geospatial datasets. This
paper describes a conditional GAN-based stochastic geospatial downscaling
method that can accommodates very high scaling factors. Compared to most
existing methods, the method can generate high-resolution accurate climate
datasets from very low-resolution inputs. More importantly, the method
explicitly considers the uncertainty inherent to the downscaling process
that tends to be ignored in existing methods. Given an input, the method
can produce a multitude of plausible high-resolution samples instead of one
single deterministic result. These samples allow for an empirical
exploration and inferences of model uncertainty and robustness. With a case
study of gridded climate datasets (wind velocity and solar irradiance), we
demonstrate the performances of the framework in downscaling tasks with
large scaling factors (up to $64\times$) and highlight the advantages of
the framework with a comprehensive comparison with commonly used and most
recent downscaling methods, including area-to-point (ATP) kriging, deep
image prior (DIP), enhanced super-resolution generative adversarial
networks (ESRGAN), physics-informed resolution-enhancing GAN (PhIRE GAN),
and an efficient diffusion model for remote sensing image super-resolution
(EDiffSR).
\end{abstract}

\begin{graphicalabstract}
\end{graphicalabstract}

\begin{highlights}

\item We describe a LAG-based framework for stochastic downscaling of gridded scientific datasets with large scaling factors, accounting for complex spatial patterns and uncertainty in the downscaling process;

\item We demonstrate the performance of the framework with large scaling factors (up to $64\times$ for wind velocity and solar irradiance data) through a comprehensive comparison with a range of commonly used downscaling methods;

\item We examine the uncertainty and robustness of the LAG-based models with a simulation-based approach and check the mass-preservation of the model outputs.

\end{highlights}


\end{frontmatter}



\section{Introduction}
\label{sec:introduction}

Accurate and high-resolution mappings of geospatial datasets are crucial for scientific research and decision-making. Taking climate data as an example, many climate models, such as general circulation models, are numerical models that mathematically describe the intricate, non-linear dynamics of the climate system and the complex interplay between climate factors \citep{edwards2011history}. Due to the inherent complexity, these numerical models are often extremely demanding for computational resources. Even high-performance computers can only operate on grids with relatively coarse spatial resolutions. Consequently, the outputs of the numerical models often fall short of capturing the essential details of spatial variations in climate variables. It highlights the need for effective downscaling methods to improve gridded climate datasets and other gridded scientific datasets in general \citep{overpeck2011climate,schumann2020need}.

The concept of downscaling is to make estimations at a finer spatial scale than that of the original datasets, aiming to enhance information and details. Within climate sciences, many climate downscaling methods have been developed and generally fall into two categories: process-based models and statistical downscaling methods \citep{hewitson1996climate, ekstrom2015appraisal}. Process-based models focus on the numerical modeling of regional or nested climate dynamics, which can also be computationally demanding. The statistical methods, on the other hand, aim to enhance the spatial details by modeling spatial patterns and leveraging the empirical relations between coarse and fine-scale climate variables. Statistical downscaling is a traditional task shared by many related disciplines. Scale is inherent to spatiotemporal data, and changing spatiotemporal scales is one of the most fundamental problems in GIScience and spatial statistics \citep{goodchild1993framework, cressie1996change}. In computer vision and machine learning, a closely related topic is image super-resolution (SR), which is to reconstruct high-resolution (HR) images from low-resolution (LR) images. In recent years, with the advancements of deep neural network-based methods \citep{Lecun2015}, a multitude of image SR methods have emerged to harness the power of deep neural network structures in representing intricate spatial patterns. The deep neural network structures, particularly convolutional neural networks (CNNs), equipped with a substantial amount of learnable weights, can effectively model and capture the complex spatial patterns embedded in LR images, and generate HR images that exhibit the minimal discrepancy when compared to LR images. These deep learning-based SR methods have shown superior performance compared to traditional methods in image SR; see \citep{yang2019deep, Wang2019d, Chauhan2023} for reviews of the recent advancements.

The deep learning-based SR methods have been successfully applied for downscaling climate and scientific datasets \citep{jiang2019edge, stengel2020adversarial, Leinonen2021, jozdani2022review, xiao2022generating, wang2023seamless, xiao2023ediffsr, xiao2024ttst}. For example, the edge-enhanced GAN (EEGAN) \citep{jiang2019edge} that includes an edge-enhancement subnetwork (EESN) designed explicitly for edge extraction and enhancement in satellite images during the super-resolution process; also the diffusion-based EDiffSR \citep{xiao2023ediffsr} and transformer-based TTST \citep{xiao2024ttst} for remote sensing image super-resolution. Despite the success, climate and scientific measurements are inherently different from natural images that most SR methods were developed for and hence pose additional challenges for downscaling methods. First, uncertainty is an unavoidable property of scientific measurements \citep{goodchild2020well, cao2022deep}, and it is vital to characterize the uncertainty and understand its impact on scientific models and decision-making \citep{postels2020quantifying}. Downscaling is a typical inverse problem \citep{Tarantola2005} that does not have a unique solution; a number of HR images can correspond to the exact same LR image, and the number of images grows exponentially as the scaling factor increases. Thus, the downscaling process inevitably introduces new uncertainty to the results, and successful downscaling methods should capture the \textit{one-to-many} mappings between LR and HR images. While significant advancements have been made in uncertainty quantification in deep learning methods \citep{ mackay1992practical, lakshminarayanan2017simple, wilson2020case, Abdar2021, rahaman2021uncertainty}, most deep learning-based downscaling methods tend to produce deterministic results, often overlooking such uncertainty. Secondly, scientific datasets often exhibit a wide variety of spatial resolutions, ranging from tens of kilometers to mere meters. Harmonizing this broad spectrum of spatial resolution requires downscaling methods that can function effectively with large scaling factors. We refer the downscaling tasks with large scaling factors (e.g., $64\times$) as \textit{extreme downscaling}. In this study, we aim to address these challenges by taking advantage of the advancements in generative adversarial networks. Furthermore, the pixel values of scientific datasets bear physical meanings of geographic regions. In the absence of additional information, downscaled results should be aligned with the original inputs. It is known as mass-preserving or pycnophylactic property in GIScience and geostatistics literature \citep{tobler1979smooth, kyriakidis2004geostatistical, kyriakidis2005geostatistical, atkinson2013downscaling}.

Within the spectrum of deep learning architectures, generative adversarial networks (GANs) \citep{goodfellow2020generative} and the variant conditional GAN (CGAN) \citep{mirza2014conditional} provide a compelling framework for stochastic image SR and other image restoration tasks \citep{ledig2017photo, lim2017enhanced, wang2018esrgan}. A GAN includes two integral components, \textit{generator} and \textit{discriminator} (or \textit{critic}), both commonly realized through neural networks. When conditional data such as LR images $\vecy$ are available, the \textit{generator} can be described as a function $G_{\vectheta}(\vecz, \vecy)$ regulated by learnable parameters $\vectheta$, where $\vecz$ is a random vector with a known probability distribution (e.g., a Gaussian distribution). The main goal of the generator is to model the high-dimensional posterior distribution of HR images $\vecx$ guided by $\vecy$, $\mathbb{P}_{\vecx\mid\vecy}$ with the help of $\vecz$, or convert $\vecz$ to $\vecg_{\vectheta} = G_{\vectheta}(\mathbf{z}, \mathbf{y})$ such that $\vecg_{\vectheta}\mid\vecy \sim \mathbb{P}_{\vecx|\vecy}$. In contrast, the critic function, $C_{\vecomega}(\vecx, \vecy)$ regulated by learnable parameters $\vecomega$, acts as an adversary to the generator to discriminate $G_{\vectheta}(\vecz, \vecy)$ from $\mathbb{P}_{\vecx\mid \vecy}$. The generator $G(\vecz, \vecy)$ and critic $C(\vecx, \vecy)$ operate in an adversary manner, undergoing simultaneous training through a two-player minimax game framework. Training a CGAN amounts to modeling the high-dimensional posterior distribution of HR images given LR images through a deep learning approach. The GAN-based models are free of the assumptions in traditional theory of statistics and probability, flexible in capturing complex spatial patterns, and yet share the generative properties of probability distributions. These properties render CGAN a suitable framework for uncertainty-aware scientific downscaling.

Recent work has emerged to take advantage of the CGAN in stochastic climate downscaling. Notable examples include \citep{Leinonen2019, Leinonen2021, Jiang2023} that use the CGANs to downscale the gridded scientific datasets and to explore the uncertainty of outcomes with generated samples. However, these CGANs often assume that pairs of LR and HR are deterministic, thereby failing to capture the one-to-many stochasticity discussed previously and possibly leading to mode collapse \citep{Yang2019a}. Moreover, these CGANs were not designed for downscaling with large scaling factors. Progress has been made recently in addressing these issues. Notably, a variant of CGAN, namely latent adversarial generator (LAG) \citep{berthelot2020creating}, was proposed to account for the stochasticity of one-to-many mappings in the downscaling process by assuming an HR image is a possibility rather than a single choice corresponding to LR images. It is achieved by learning a Gaussian distributed latent variable representing the distances between prediction and ground truth. To support super-resolution with large scaling factors, a progressive training strategy for GAN (ProGAN \citep{karras2017progressive}) was developed. Starting with a low image scaling factor, the model captures incrementally fine details by adding new layers as the training progresses and the factor increases. This strategy has been proven to not only expedite the training process and enhance stability but also better capture variations across multiple scales \citep{karras2017progressive}.

In this study, we describe a stochastic downscaling framework for extreme downscaling of gridded scientific datasets, leveraging both ProGAN \citep{karras2017progressive} and LAG \citep{berthelot2020creating}. The framework enjoys the advantages of both sides and can efficiently address the previously discussed challenges in scientific downscaling: ProGAN for large scaling factors and LAG for uncertainty modeling. With case studies of gridded climate datasets (wind velocity and solar irradiance), we demonstrate the performances of the framework in downscaling tasks with large scaling factors (e.g., $64\times$) and highlight the advantages of the framework with a comprehensive comparison with latest commonly used downscaling methods, including area-to-point (ATP) kriging \citep{kyriakidis2004geostatistical}, deep image prior (DIP) \citep{ulyanov2018deep}, enhanced super-resolution generative adversarial networks (ESRGAN) \citep{wang2018esrgan}, physics-informed resolution-enhancing GAN (PhIRE GAN) \citep{stengel2020adversarial}, and an efficient diffusion model for remote sensing image super-resolution (EDiffSR) \citep{xiao2023ediffsr}. The mass-preservation and the uncertainty space of the trained LAG-based models are also examined.

The contributions of this work are summarized as follows:

\begin{enumerate}
\item We describe a LAG-based framework for stochastic downscaling of gridded scientific datasets with large scaling factors, accounting for complex spatial patterns and uncertainty in the downscaling process;
\item We demonstrate the performance of the framework with large scaling factors (up to $64\times$ for wind velocity and solar irradiance data) through a comprehensive comparison with a range of commonly used downscaling methods;
\item We examine the uncertainty and robustness of the LAG-based models with a simulation-based approach and check the mass-preservation of the model outputs.
\end{enumerate}

The rest of this article is organized as follows: Section \ref{sec:related_work} introduces background and related work to this study. Section \ref{sec:methodology} presents the methodology of the described LAG-based downscaling framework. Section \ref{sec:experiments} provides details of the experimental design for a comprehensive performance comparison applied to gridded climate datasets, including the datasets we are using and the training configuration of our framework and other comparison models. Section \ref{sec:results} analyzes the experimental results from various aspects, including accuracy statistics, visual analysis, mass preservation, and uncertainty characterization. The advantages of the LAG-based framework are highlighted in this section. Section \ref{sec:conclusion} concludes this paper and discusses future work.

\section{Related work} \label{sec:related_work}

\subsection{Geostatistical methods}

Downscaling images or gridded datasets can be taken as a particular case of the traditional change of spatial support problem (COSP) in spatial statistics \citep{cressie1996change, gotway2002combining} or modifiable areal unit problem in GIScience \citep{Openshaw1979, gotway2002combining,kyriakidis2004geostatistical, Goodchild2022}. Numerous frameworks have been established within statistics and GIScience to address these problems, many of which have been applied for image downscaling. For example, in geostatistics, area-to-point kriging (ATPK) \citep{kyriakidis2004geostatistical, kyriakidis2005geostatistical} was developed to make predictions at points from areal aggregations and has been widely adopted for downscaling satellite images \citep{Pardo2006, atkinson2008downscaling, wang2015downscaling,tang2015downscaling} and gridded scientific data \citep{zhang2012geostatistical}.

Kriging methods enjoy many advantages. Grounded in statistics theory, they provide measures of reliability or uncertainty for predictions \citep{kyriakidis2004geostatistical,kyriakidis2005geostatistical}. They can yield coherent and mass-preserving predictions that are important for scientific modeling. It means the statistics of disaggregated pixels (e.g., average) in downscaled images are aligned with the values of the associated coarse pixels in the original images. Furthermore, kriging methods provide effective tools, such as variograms, to examine the spatial patterns inherent in spatial datasets. Despite the advantages, as in other traditional statistical methods, kriging methods were developed under statistical assumptions that can be difficult to satisfy in practice. The reliance on two-point statistics, such as covariance functions, limits the performance in modeling complex spatial patterns \citep{cao2022deep}.

\subsection{Deep learning methods}

\subsubsection{GAN-based methods and others}
As one of the most representative structures in deep learning, the GAN-based models, CGANs in particular, have demonstrated successes in image super-resolution and the potential to address the challenges in climate downscaling. Unlike ordinary GANs, CGANs feed the generator with additional data, such as an LR image, to guide the generation of HR outcomes. See \citep{Singla2022} for a recent review of such methods. Among these methods, super-resolution GAN (SRGAN)\citep{ledig2017photo} and enhanced super-resolution GAN (ESRGAN) \citep{wang2018esrgan} are the most commonly used \citep{Kurinchi-Vendhan2021, yang2019deep}. SRGAN integrated a ResNet-based deep neural network architecture (SRResNet) into the GAN framework. SRGAN designed a perceptual loss function to ensure high perceptual quality and peak signal-to-noise ratios in the outcomes. The perceptual loss consists of an adversarial loss and a content loss based on VGG models \citep{simonyan2014very}: the former encourages the output to align with the manifold of training images, while the latter maintains the perceptual similarity with input images. ESRGAN enhances the SRGAN in network structure and loss functions to further reduce the artifacts of the generated images \citep{wang2018esrgan}. These GAN-based methods have been widely applied for downscaling scientific datasets, including remote sensing imagery and meteorological measurements \citep{Kurinchi-Vendhan2021, stengel2020adversarial, Leinonen2019}.

In addition to GAN-based methods, CNN and the most recent transformer and diffusion models also made notable progress in image SR. For instance, the enhanced deep residual network (EDSR) \citep{lim2017enhanced}, an efficient and lightweight model derived from SRResNet in SRGAN, improves performance and simplicity by removing the batch normalization and the ending activation layers in the residual block. The SR3 \citep{saharia2022image} method employs denoising diffusion probabilistic models (DDPM) \citep{ho2020denoising} to perform image super-resolution through a stochastic iterative denoising process. The hybrid attention transformer (HAT) \citep{chen2023activating} enhances single image super-resolution quality by integrating channel attention and self-attention mechanisms to activate more pixels. These methods have gradually been applied in scientific dataset downscaling, especially for remote sensing imagery \citep{lei2021transformer, he2022dster, chen2023msdformer}. For example, the EDiffSR \citep{xiao2023ediffsr} provides an efficient way to downscale remote sensing images with better perceptual quality.

\subsubsection{Stochastic downscaling}
Most of the aforementioned methods were designed for a single deterministic HR output. Notable progresses were made to introduce stochasticity to the CGAN-based methods that allow generating a set of plausible HR images for a given LR image \citep{Leinonen2019, Bahat2020}. The stochasticity is important to assess the uncertainty and reliability associated with model output. Several methods were developed to allow stochasticity in image SR. For instance, the uncertainty-aware GAN (UGAN) designed an uncertainty-aware adversarial training strategy to constrain the training process according to the uncertainty in downscaling results and help the model focus more on the high-variance area \citep{ma2024uncertainty}. The photo upsampling via latent space exploration (PULSE) method explores the latent spaces of the pre-trained GANs and samples the possible HR images that are upscalable to the given LR images \citep{Menon2020}. In unsupervised learning, deep image prior (DIP) \citep{ulyanov2018deep} showed that the neural network structure could serve as an image prior that can be stochastically sampled from and achieve comparable performances with supervised methods. Many of these methods, however, treated the $(\vecx, \vecy)$ pair as deterministic and attempted to estimate the posterior distribution $\mathbb{P}_{\vecx|\vecy}$ using only a single sample $\vecx$ for each $\vecy$. Consequently, these approaches may result in mode collapse \citep{Yang2019a, goodfellow2020generative} and are unable to effectively consider the inherent uncertainty arising from the one-to-many mappings involved in the downscaling process. In contrast, LAG was recently proposed to assume an HR image is one sample from the manifold of LR images and introduced a Gaussian-like latent space to quantify the distances between LR and HR images \citep{berthelot2020creating}.

\subsubsection{Extreme downscaling}
Applying GAN-based methods for downscaling tasks with large scaling factors (e.g., $64\times$) can lead to unstable training processes and mode collapses \citep{arjovsky2017towards, arjovsky2017wasserstein}. When mode collapse happens, the generator produces a similar or a limited variety of outputs. One effective strategy to mitigate this is to break down the training process into multiple steps. For instance, PhIRE GAN adopts deep CNNs and SRGAN for a two-step SR \citep{stengel2020adversarial}. The SR model in the first step produces images from LR to medium resolution (MR), and the SR model in the second step is responsible for MR to HR. With the two-step process, the PhIRE GAN achieved up to $50\times$ resolution enhancement for the Global Climate Model (GCM) outputs while preserving physically relevant characteristics. To streamline the process, ProGAN was recently developed to train the GAN models progressively, which facilitates the extreme downscaling \citep{karras2017progressive}.

\section{Methodology}\label{sec:methodology}

Given an LR image $\vecy$, our goal is to sample from $\mathbb{P}_{\vecx\mid\vecy}$, the high dimensional posterior distribution of HR images $\vecx$ that are possible downscaling outcomes of $\vecy$. With the help of a random vector $\vecz$, CGAN achieved this by mapping a pair of $(\vecz,\vecy)$ to $\vecg_{\vectheta}=G_{\vectheta}(\vecz,\vecy)$ such that $\vecg_{\vectheta}\mid \vecy \sim \mathbb{P}_{\vecx\mid\vecy}$. Hence, sampling from $\vecz$ effectively corresponds to sampling from $\mathbb{P}_{\vecx|\vecy}$. To avoid mode collapse and to account for the stochasticity in the one-to-many mappings in the downscaling process discussed previously, we adopt a method inspired by LAG \citep{berthelot2020creating}. As in other CGAN methods, LAG employs a simple Gaussian random vector $\vecz \in R^N, \vecz\sim\mathcal{N}(\veczero,\vecone)$. Within a LAG framework, it further assumes that possible HR images $\vecx$ corresponding to an LR image $\vecy$ follows a latent Gaussian-like distribution centered around LR images $\mathcal{P}_{\vecz, \vecy}(\vecx)$, and that the most possible HR image should occur in the distribution center when $\vecz = 0$. The assumption explicitly considers the stochastic links between $\vecx$ and $\vecy$, and can be implemented by introducing an additional regularization term in the generator loss, which will be detailed below.

\subsection{Loss functions}
To accurately model the posterior distribution $\mathbb{P}_{\vecx|\vecy}$ with the generator $G_{\vectheta}(\vecz,\vecy)$, it is important to seek optimal generator and critic parameters that can minimize the discrepancy between $\mathbb{P}_{\vecx|\vecy}$ and $\mathbb{P}_{\vecg_{\vectheta}|\vecy}$. In GAN literature, different loss functions were defined to quantify the discrepancy while balancing between perceptual quality and distortion \citep{Blau2018}. As in LAG, we adopt the Wasserstein GAN (WGAN) loss with gradient penalty \citep{arjovsky2017wasserstein, Gulrajani2017} in this study. To capture the proposed stochastic relationship between $\vecx$ and $\vecy$, an additional regularization term  was introduced to the WGAN loss function:

\begin{equation}\label{eq:loss}
\mathcal{L}(\mathbb{P}_{\vecx|\vecy}, \mathbb{P}_{\vecg_{\vectheta}|\vecy}|\vectheta, \vecomega) = \mathcal{L}_{wgan}(\vecx,\vecy) + \lambda_{gp} \cdot \mathcal{L}_{gp}(\vecx, \vecy) + \lambda \cdot \mathcal{L}_{center}(\vecx, \vecy)
\end{equation}

\noindent where $\mathcal{L}_{wgan}(\cdot, \cdot)$ indicates the regular WGAN loss \citep{arjovsky2017wasserstein}, $\mathcal{L}_{gp}(\cdot, \cdot)$ is the gradient penalty to enforce the 1-Lipschitz constraints on the critic function \citep{Gulrajani2017}, $\lambda_{gp}$ and $\lambda$ are the regularization coefficients. The term $\mathcal{L}_{center}(\cdot, \cdot)$ is the introduced regularization term penalizing deviations of $\vecx$ from the center of the latent distribution $\mathcal{P}$, which can be written as:

\begin{equation}\label{eq:loss_G_center}
\mathcal{L}_{center}(\vecx,\vecy) = \mathbb{E}_{\vecx,\vecy} [||P(\vecx, \vecy)-P(G_{\vectheta}(\veczero|\vecy), \vecy)||_{2}^2]
\end{equation}

\noindent The function $P(\vecx,\vecy)$, as defined in Eq.~\ref{eq:loss_G_center}, maps $\vecx$ into a location on the latent distribution $\mathcal{P}$. This function, implemented in a deep neural network architecture, is designed to capture the complex spatial patterns in $\vecx$. The notation $G_{\vectheta}(\veczero, \vecy)$ represents the center of the latent distribution. The logic of the loss function $\mathcal{L}_{center}(\vecx,\vecy)$ is the ground truth (in the training process) or the most likely image (in the sampling process) should be close to the distribution center. It allows for an HR image $\vecx$ to be taken as just one possible outcome of downscaled $\vecy$, instead of a single, deterministic result. In doing so, it accounts for the intrinsic variability in the downscaling process and contributes to mitigating mode collapse by allowing multiple potential downscaling solutions. From $P(\vecx, \vecy)$, the ciritic function can be derived, i.e., $C_{\vecomega}(\vecx, \vecy)=F(P(\vecx, \vecy))$, where $F(\cdot)$ maps the locations in $\mathcal{P}$ into a scalar critic score. For the mapping function $F(\cdot)$, simple mathematical operators are applicable. In this study, the mean operator is employed.

To summarize, the loss function in Eq.~\ref{eq:loss} can be expanded into Eq.~\ref{eq:loss_details}, and as in CGAN, the generator $G_{\vectheta}$ and critic $C_{\vecomega}$ are trained via a two-player minimax game:

\begin{equation}\label{eq:loss_details}
  \begin{split}
  {\displaystyle \min_{\vectheta} \max_{\vecomega}} \;
  & \mathbb{E}_{\vecx}[C_{\vecomega}(\vecx, \vecy)] - \mathbb{E}_{\vecz}[C_{\vecomega}(G_{\vectheta}(\vecz,\vecy), \vecy)] \\
  & + \lambda_{gp} \cdot \mathbb{E}_{\hat{\vecx},
  \vecy}\left[\left(\left\|\nabla_{\hat{\vecx}} C_\omega(\hat{\vecx},
  \vecy)\right\|_2-1\right)^2\right] \\
  & + \lambda \cdot \mathbb{E}_{\vecx,\vecy} [||P(\vecx, \vecy)-P(G_{\vectheta}(\veczero|\vecy), \vecy)||_{2}^2]
  \end{split}
\end{equation}

\noindent where $\hat{\vecx}$ indicates a uniform sample on the straight lines between $\vecx$ and $\vecg_{\vectheta}$. The generator function $G_{\vectheta}$ and the critic function $C_{\vecomega}$ are both implemented in deep neural networks, as will be detailed in the following subsection.

\subsection{Neural networks structures}

\begin{figure*}
  \centering
  \includegraphics[width=\columnwidth]{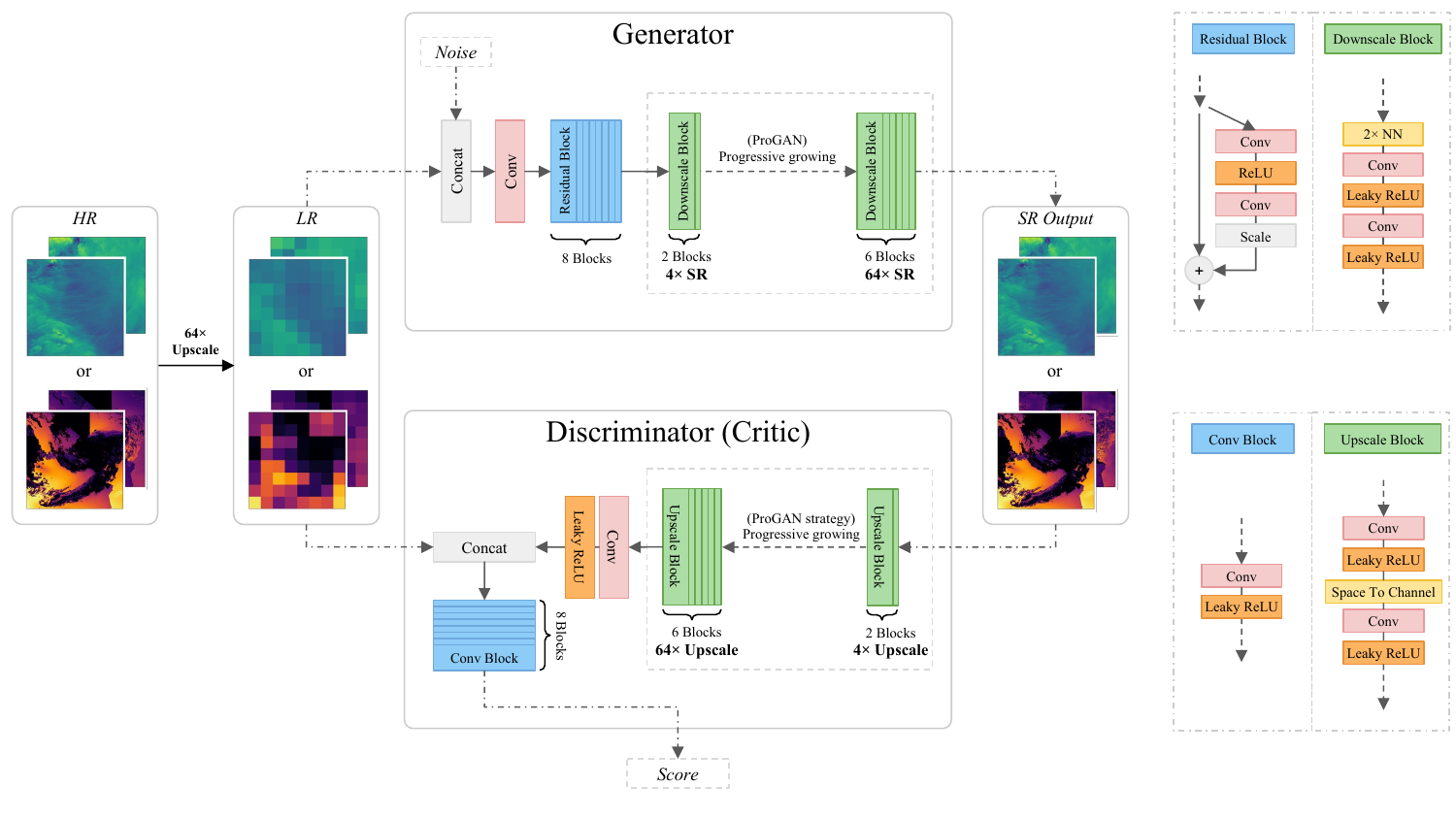}
  \caption{Structures of the generator and critic in the
    described LAG-based downscaling framework.}
  \label{fig:lag_diagram}
\end{figure*}

Fig.~\ref{fig:lag_diagram} presents the detailed structure of the LAG-based generator and critic. We adopt the residual network from EDSR \citep{lim2017enhanced} as a basic downscale block for its lightweight and efficiency, then grow it under the LAG framework using the progressive training strategy of ProGAN \citep{karras2017progressive} to achieve a larger scaling factor. Each downscale block in the generator consists of a $2\times$ nearest neighbor interpolation layer and two convolution layers with the Leaky ReLU activation function separately. The critic uses a similar structure as the generator but in reverse order. The nearest neighbor interpolation layer in the downscale block of the generator is replaced by a new layer, which moves space to channels to turn downscale blocks into upscale blocks.

With the progressive training strategy of ProGAN, new downscale and upscale blocks can be systematically added to the generator and critic networks, respectively, to ensure that all layers remain trainable and stable across the entire training process. It is worth noting that there is an extra transition phase before the training progresses to smoothly incorporate the newly added layers by increasing their weights from 0 to 1. The transition phase is irreplaceable in avoiding sudden changes in image resolution; see \citep{karras2017progressive} for more information.

\section{Experimental design}\label{sec:experiments}

To highlight the advantages of the described LAG-based downscaling framework for stochastic extreme downscaling, we applied the method to downscale gridded climate datasets and compared the performances with the current state-of-the-art methods from both quantitative and qualitative perspectives. The characteristics and advantages of the LAG-based model, such as model uncertainty and mass-preservation, were also examined. All the codes (in Python) and pre-trained weights are available at \url{https://github.com/LiGuiye/LAG_Climate}.

\subsection{Datasets}

The datasets we used are gridded data of wind velocity and solar irradiance covering the continental United States. The datasets were obtained from the National Renewable Energy Laboratory's (NREL's) Wind Integration National Database (WIND) Toolkit \citep{draxl2015overview} and the National Solar Radiation Database (NSRDB) \citep{sengupta2018national}. Examples of the two gridded datasets are given in Fig.~\ref{fig:testSamples_Wind_4X} and \ref{fig:testSamples_Solar_4X}. As can be observed from the data, the solar irradiance dataset exhibits more complex patterns than the wind velocity dataset. Both datasets have been used as a benchmark for evaluating downscaling methods \citep{stengel2020adversarial, Kurinchi-Vendhan2021}.

The WIND Toolkit provides $4$-hourly data on a uniform $2$-km grid. We chose the $100$-meter height wind speed and direction data points and converted them into easterly ($u$) and northerly ($v$) wind velocities as two different image channels. Data points were collected from $2007$ to $2014$ within three randomly selected $1,024$ square kilometer areas of the continental United States. The NSRDB contains half-hourly measurements for Direct Normal Irradiance (DNI) and Diffuse Horizontal Irradiance (DHI) at $4$-km spatial resolution in latitude and longitude. DNI and DHI are the amounts of solar radiation received per unit area that reaches the earth directly and indirectly from the sun. We sampled the data hourly during daylight hours (from $6$ AM to $6$ PM) from $2007$ to $2014$ for data quality and diversity.

\begin{table}
  \caption{Details of the training and testing datasets.}
  \label{tab:datasetInfo}
  \centering
  \begin{tabular}{@{}lrr@{}}
    \toprule
    Dataset             & Wind velocity & Solar irradiance \\\midrule
    Years               & $2007-2014$   & $2007-2014$      \\
    Image size          & $512\times512$& $512\times512$  \\
    Training size       & $46,026$      & $30,056$         \\
    Testing size        & $5,605$       & $3,648$          \\
    Spatial resolution  & $2$ km        & $4$ km           \\
    Temporal resolution & $4$-hourly    & hourly           \\
    \bottomrule
  \end{tabular}
\end{table}

The chosen wind velocity and solar irradiance data were mapped to gridded images (with image size $512\times512$) and chronologically split into training and testing datasets. The datasets from $2007$ to $2013$ were used for training, and the ones from $2014$ were used for testing. The training sets of wind velocity and solar irradiance data contain $46,026$ and $30,056$ images, respectively. The test sets include $5,605$ wind velocity grids and $3,648$ solar irradiance grids. Table \ref{tab:datasetInfo} summarizes the details of the datasets.

\subsection{Training details}\label{sec:trainingDetails}
The HR images in the training datasets (with image size $512\times512$) were first upscaled to LR image sizes ($8\times8$) using average pooling. The upscaled (LR) images ($\vecy$) and original HR images ($\vecx$) were for training the model. We employed an Adam optimizer \citep{kingma2014adam} with $\alpha_1 = 0$, $\alpha_2 = 0.99$, and $\epsilon=1e^{-8}$ for optimization and an exponential moving average with a decay of $0.999$ was used for the weights of the generator for better visual effects. The generator and critic used the same learning rate, and we trained the critic functions more times than the generator for better results.

Models for wind velocity and solar irradiance data were trained separately using $1$ NVIDIA A100 GPU. For the wind velocity data, we trained the model $30$ epochs for each transition and stabilization phase at different resolutions with a batch size of $16$ and a learning rate of $2 \times 10^{-3}$. For the solar irradiance data, we found that smaller batch sizes helped to improve the model performance during the evaluation process, possibly due to the heterogeneity of the images. We used a batch size of $1$ and a learning rate of $4 \times 10^{-3}$ to train the model for $15$ epochs at each transition and stabilization phase.

\subsection{Accuracy metrics}
Similar to the training process, during testing, we upscaled the HR images in the testing dataset as the model inputs, the original HR images were used as references. We used relative Mean Squared Error (MSE) and Sliced Wasserstein Distance (SWD) \citep{rabin2012wasserstein} to measure the discrepancy between the generated images of the LAG-based method and reference images. The relative MSE, MSE divided by the average value in ground truth images, measures the pixel-wise accuracy, while the SWD, a variant of Wasserstein Distance or Earth-Mover distance \citep{rubner1998metric, rubner2000earth}, measures the perceptual accuracy. For SWD, we adopted the same calculation methods used in ProGAN \citep{karras2017progressive}, first normalizing each channel and then computing the mean value of SWDs across all channels.

Furthermore, we used semivariograms to evaluate the performance of each method in reproducing spatial patterns. As mentioned previously, the semivariogram is a commonly used geostatistical tool to measure the spatial variations of spatial measurements \citep{chiles1999geostatistics,sain2011spatial}. Despite being a two-point statistic, it provides further insights into how measurements vary over the space. Datasets with comparable spatial patterns are expected to have closely aligned semivariograms, reflecting similar spatial continuity and variability.

\subsection{Model comparisons}
For a comprehensive comparison, we selected five different types of models for image super-resolution, including geostatistical methods, unsupervised methods, GAN-based, and diffusion-based methods. These methods include ATPK \citep{kyriakidis2004geostatistical,kyriakidis2005geostatistical}, DIP \citep{ulyanov2018deep}, ESRGAN \citep{wang2018esrgan}, PhIRE GAN \citep{stengel2020adversarial}, and EDiffSR \citep{xiao2023ediffsr}. The performance of these methods was compared at regular ($4\times$ and $8\times$) and large scaling factors ($64\times$). PhIRE GAN uses a two-step process and can perform $50\times$ SR for the wind velocity and $25\times$ SR for the solar irradiance data. The pre-trained weights from \url{https://github.com/NREL/PhIRE} were used to generate test samples for comparison \citep{stengel2020adversarialCode}. All other SR models are retrained with the abovementioned datasets using official settings to ensure a fair comparison.

\section{Results}
\label{sec:results}

\subsection{Accuracy statistics}\label{sec:accuracy}

\begin{table}
  \begin{center}
    \caption{Accuracy statistics comparison of generation performance of different models at \textbf{regular scaling factors}. The best and second best metrics at specific scaling factor are shown in {\color{red}\textbf{red bold}} and {\color{blue}\textbf{blue bold}}, respectively.}
    \label{tab:metricsTable}
    \resizebox{\textwidth}{!}{
    \begin{tabular}{rccccccc}\toprule
      \multicolumn{2}{c}{\textbf{Wind velocity}} & ATPK & DIP & ESRGAN & EDiffSR & LAG$^-$ & LAG \\
      \midrule
      \multirow{2}{*}{relative MSE ($\downarrow$) } & $4\times$ & 0.505 & 0.918 & {\color{blue}$\mathbf{0.391}$} & 0.427 & $0.400$ & {\color{red}$\mathbf{0.278}$}\\
      & $8\times$ & {\color{blue}$\mathbf{0.660}$} & 1.885 & 1.591 & 0.668 & 0.937 & {\color{red}$\mathbf{0.460}$}\\
      \multirow{2}{*}{SWD ($\downarrow$) } & $4\times$ & 0.167 & 0.157 & 0.145 & 0.146 & {\color{blue}$\mathbf{0.138}$} & {\color{red}$\mathbf{0.124}$}\\
      & $8\times$ & 0.170 & 0.167 & 0.192 & 0.142 & {\color{blue}$\mathbf{0.138}$} & {\color{red}$\mathbf{0.121}$}\\
      \midrule

      \multicolumn{2}{c}{\textbf{Solar irradiance}} & ATPK & DIP & ESRGAN & EDiffSR & LAG$^-$ & LAG \\\midrule
      \multirow{2}{*}{relative MSE ($\downarrow$) } & $4\times$ & 0.144 & {\color{blue}$\mathbf{0.089}$} & 0.138 & 0.137 & 0.096 & {\color{red}$\mathbf{0.081}$}\\
      & $8\times$ & 0.171 & {\color{red}$\mathbf{0.123}$} & 0.475 & 0.216 & 0.154 & {\color{blue}$\mathbf{0.128}$}\\
      \multirow{2}{*}{SWD ($\downarrow$) } & $4\times$ & 0.199 & 0.184 & 0.197 & {\color{blue}$\mathbf{0.179}$} & 0.187 & {\color{red}$\mathbf{0.157}$}\\
      & $8\times$ & 0.200 & 0.192 & 0.231 & {\color{blue}$\mathbf{0.173}$} & 0.196 & {\color{red}$\mathbf{0.151}$}\\\bottomrule
    \end{tabular}
    }
  \end{center}
\end{table}

We applied the trained models to downscale the upscaled wind velocity and solar irradiance test datasets and collected the accuracy statistics by comparing the downscaled results with the reference data. Table \ref{tab:metricsTable} shows the median relative MSE and SWD values of the compared methods at regular scaling factors ($4\times$ and $8\times$). Smaller values indicate better performances. The best and second best metrics at specific scaling factors and datasets are shown in red bold and blue bold, respectively.

The model labeled LAG$^-$ in Table \ref{tab:metricsTable} indicates the one without $\mathcal{L}_{center}$ in the loss function as an ablation study. Please see Eq.\ref{eq:loss_G_center} for the regularization term, $\mathcal{L}_{center}(\vecx,\vecy)$, and Eq.\ref{eq:loss_details} for the integral loss function of the training process. Models under the LAG framework achieved the best prediction accuracy for wind velocity and solar irradiance data at regular downscaling scales ($4\times$ and $8\times$), considering pixel and perceptual accuracy. Besides, for solar irradiance data with more complex patterns, the LAG with additional specially designed items in loss function has more significant advantages.

\begin{table}
  \begin{center}
    \caption{Accuracy statistics comparison of generation performance of different models at \textbf{large scaling factors}. The best metrics at specific scaling factor are shown in {\color{red}\textbf{red bold}}.}
    \label{tab:metricsTable2}
    \begin{tabular}{cccc}\toprule
      \multicolumn{2}{c}{\textbf{Wind velocity}} & SWD ($\downarrow$) & LPIPS ($\downarrow$) \\\midrule
      PhIRE GAN & $50\times$ & 0.162 & 0.611 \\
      LAG & $64\times$ & {\color{red}$\mathbf{0.107}$} & {\color{red}$\mathbf{0.409}$}\\\bottomrule
      \multicolumn{2}{c}{\textbf{Solar irradiance}} & SWD ($\downarrow$) & LPIPS ($\downarrow$) \\\midrule
      PhIRE GAN & $25\times$ & 0.136 & 0.696 \\
      LAG & $64\times$ & {\color{red}$\mathbf{0.123}$} & {\color{red}$\mathbf{0.573}$}\\\bottomrule
    \end{tabular}
  \end{center}
\end{table}

In addition, LAG remains stable and outperforms PhIRE GAN, designed explicitly for large-scale climate data super-resolution, at large scaling factors ($25\times$, $50\times$, and $64\times$) in Table \ref{tab:metricsTable2}. As the pixel-wise MSE is already included in the content loss of PhIRE GAN, we adopt SWD and Learned Perceptual Image Patch Similarity (LPIPS) \citep{zhang2018unreasonable} as comparison metrics regarding generation quality for fairness. Lower SWD and LPIPS indicate better results.

\subsection{Visual analysis}
We randomly selected a sample from both the wind velocity and solar irradiance test datasets (both with two channels), respectively, and compared the performances of different methods in reproducing the spatial patterns in the reference data.

\begin{figure}
  \centering
  \includegraphics[width=\columnwidth]{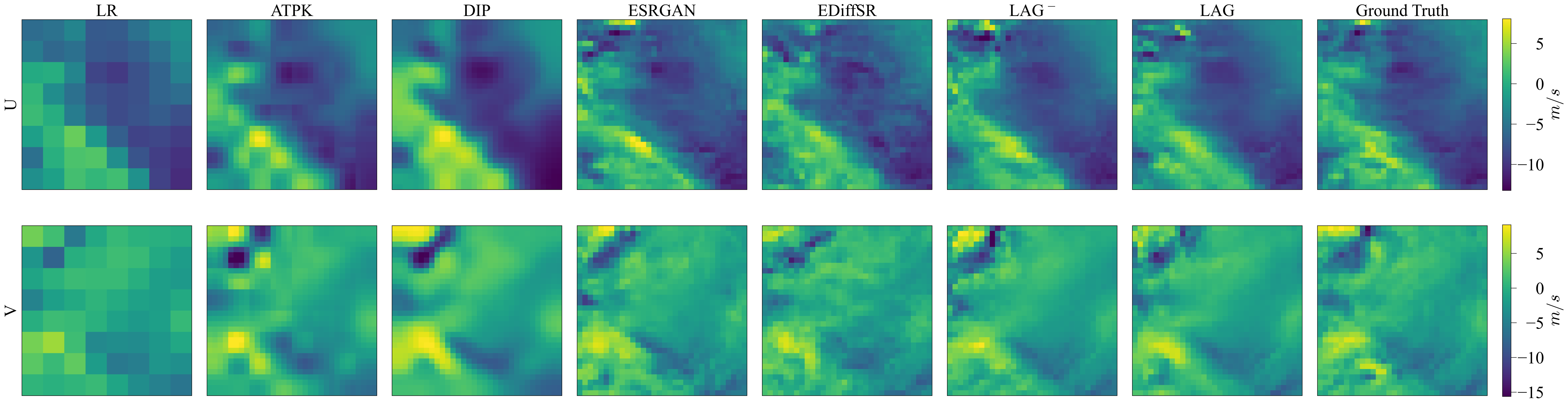}
  \caption{$4\times$ SR results generated by different methods for two channels (\textit{U} and \textit{V}) of a randomly sampled LR image from the wind velocity test dataset. Zoom in for better observation.}
  \label{fig:testSamples_Wind_4X}
\end{figure}

\begin{figure}
  \centering
  \includegraphics[width=\columnwidth]{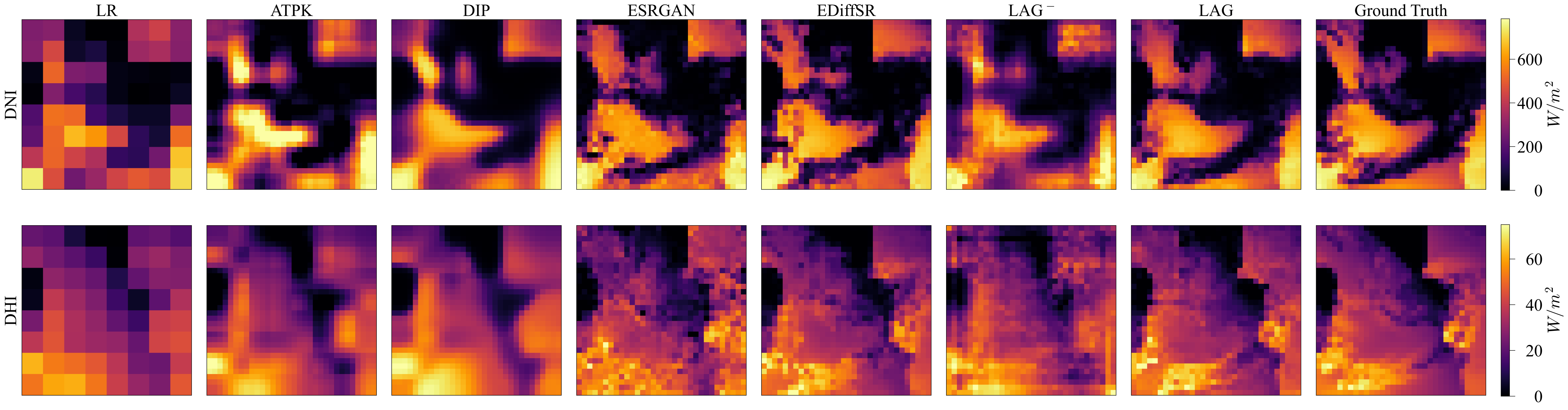}
  \caption{$4\times$ SR results generated by different methods for two channels (\textit{DNI} and \textit{DHI}) of a randomly sampled LR image from the solar radiance test dataset. Zoom in for better observation.}
  \label{fig:testSamples_Solar_4X}
\end{figure}

\begin{figure}[hbt!]
  \centering
  \subfloat[]{\includegraphics[width=.5\columnwidth]{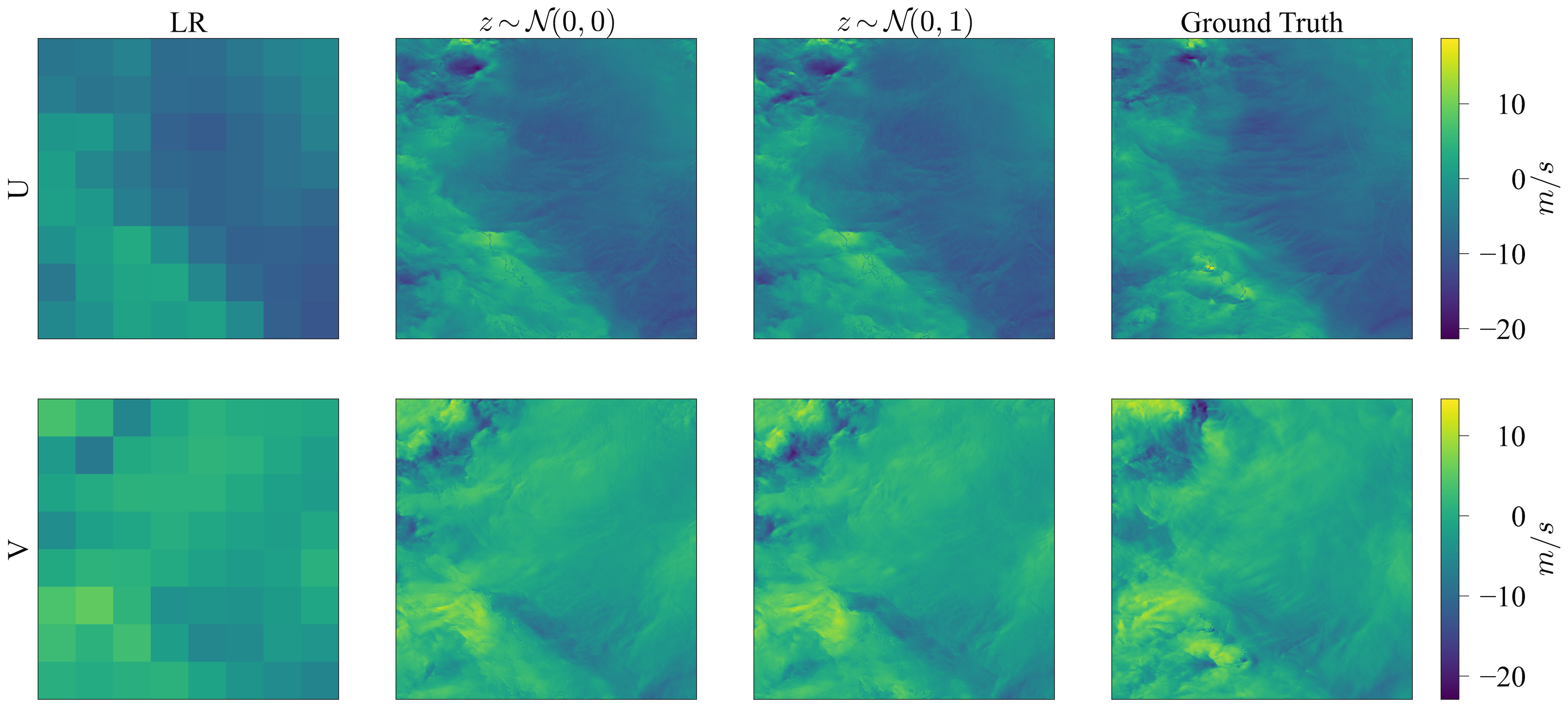}%
  \label{fig:testSamples_Wind_64X}}
  \subfloat[]{\includegraphics[width=.5\columnwidth]{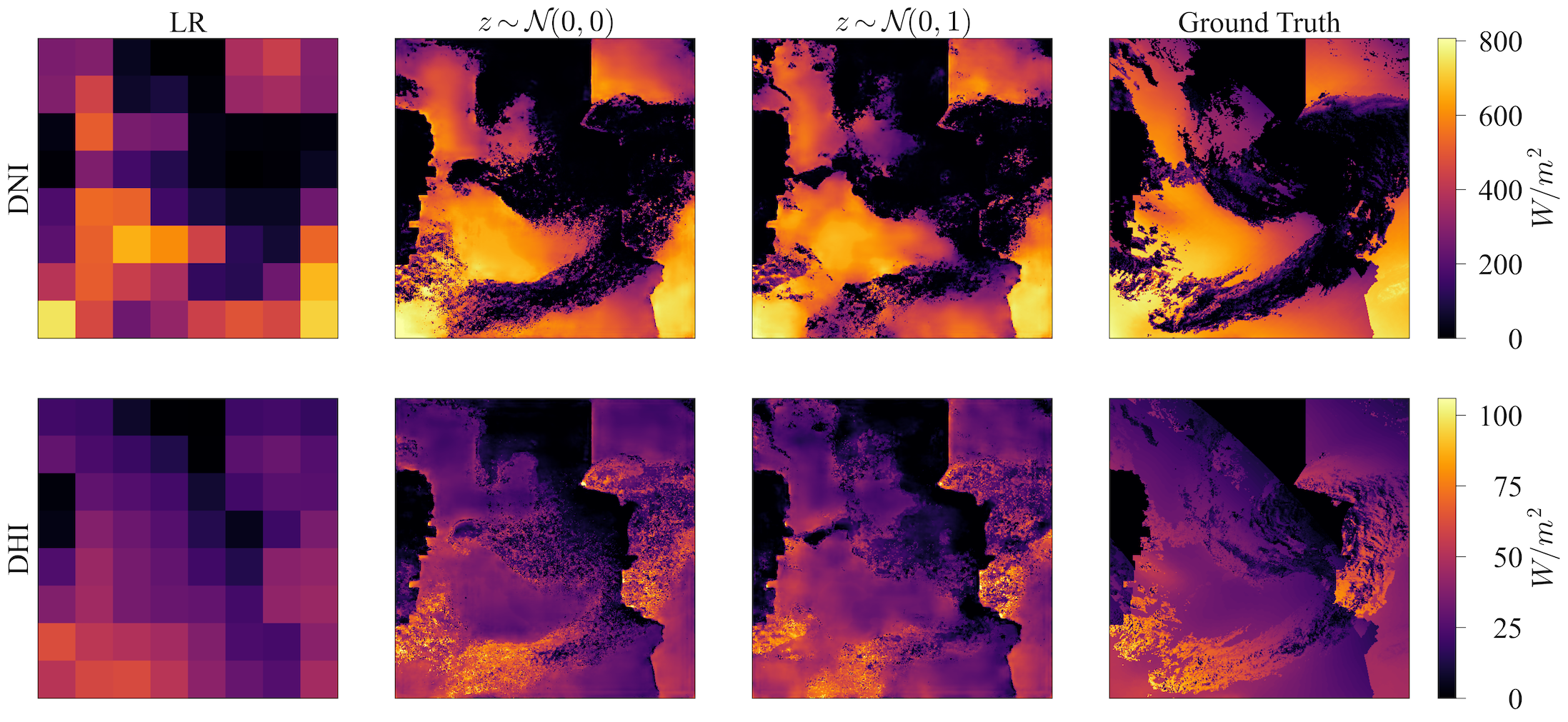}%
  \label{fig:testSamples_Solar_64X}}

  \caption{$64\times$ SR results generated by the LAG-based downscaling framework for a randomly sampled LR image from (a) wind velocity and (b) solar irradiance test datasets. For each panel, the leftmost column shows the input LR images at two channels (with image size $8\times8$), the rightmost column shows the corresponding HR ground truth images (with image size $512\times512$), and the columns in between shows results of LAG. Zoom in for better observation.}

  \label{fig:testSamples_Wind_Solar_64X}
\end{figure}

Figs.~\ref{fig:testSamples_Wind_4X} and \ref{fig:testSamples_Solar_4X} show the $4\times$ SR results generated by the compared methods for the wind velocity and solar irradiance datasets, respectively. The leftmost column indicates the input LR images (with image size $8\times8$) randomly sampled from the test set, while the rightmost column shows the corresponding HR ground truth images (with image size $32\times32$). The results of different methods are shown in the columns in between (all with image size $32\times32$). As one can see, the GAN-based and diffusion-based methods, including ESRGAN, EDiffSR, LAG$^-$, and LAG, generated much closer results to the ground truth. In contrast, ATPK and DIP resulted in overly smoothed images with losses of fine details. It is expected considering that DIP is unsupervised and ATPK only considers two-point statistics. While the ESRGAN, EDiffSR, LAG$^-$, and LAG results are visually comparable for wind velocity images (Fig.~\ref{fig:testSamples_Wind_4X}), LAG produced the overall best results for the solar irradiance data with more heterogeneity in spatial patterns (Fig.~\ref{fig:testSamples_Solar_4X}). These are consistent with the accuracy statistics in Section \ref{sec:accuracy}.

Fig.~\ref{fig:testSamples_Wind_Solar_64X} shows the $64\times$ LAG SR results for the wind velocity and solar irradiance data. The LAG method can effectively reproduce impressively fine details in the reference images even at this significant scaling factor. We have included more test results at different scaling factors in the Supplementary Material.

\begin{figure}
  \centering

  \subfloat[]{\includegraphics[width=.5\columnwidth]{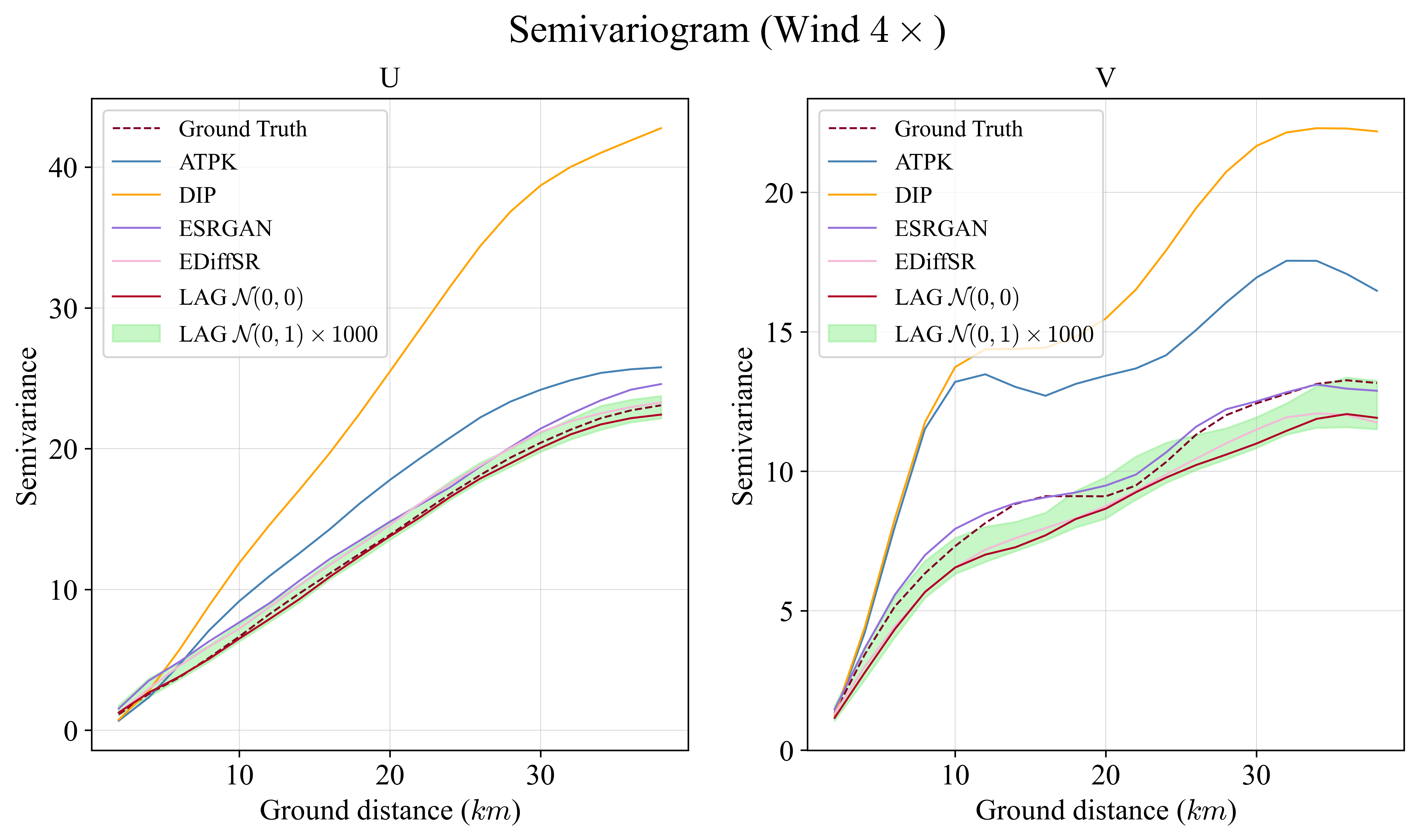}%
    \label{fig:semivariogram_Wind4X}}
  \subfloat[]{\includegraphics[width=.5\columnwidth]{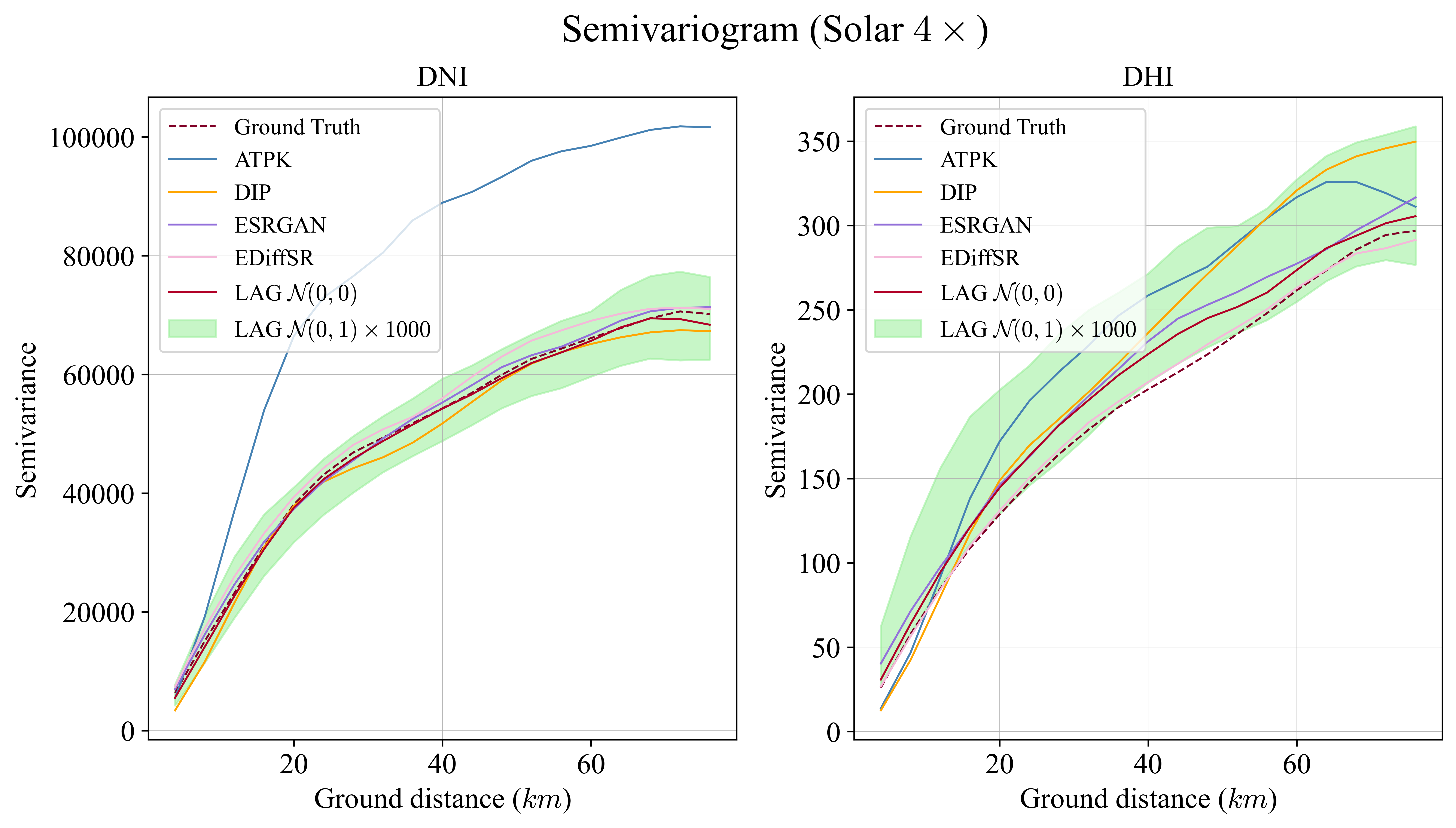}%
    \label{fig:semivariogram_Solar4X}}

  \caption{Semivariograms of the $4\times$ SR results generated by different models for a randomly sampled LR image from (a) wind velocity and (b) solar irradiance test datasets. Images with similar spatial patterns should have close lines. Zoom in for better observation.}

  \label{fig:semivariogram_Wind_Solar_4X}
\end{figure}

\begin{figure}
  \centering

  \subfloat[]{\includegraphics[width=.5\columnwidth]{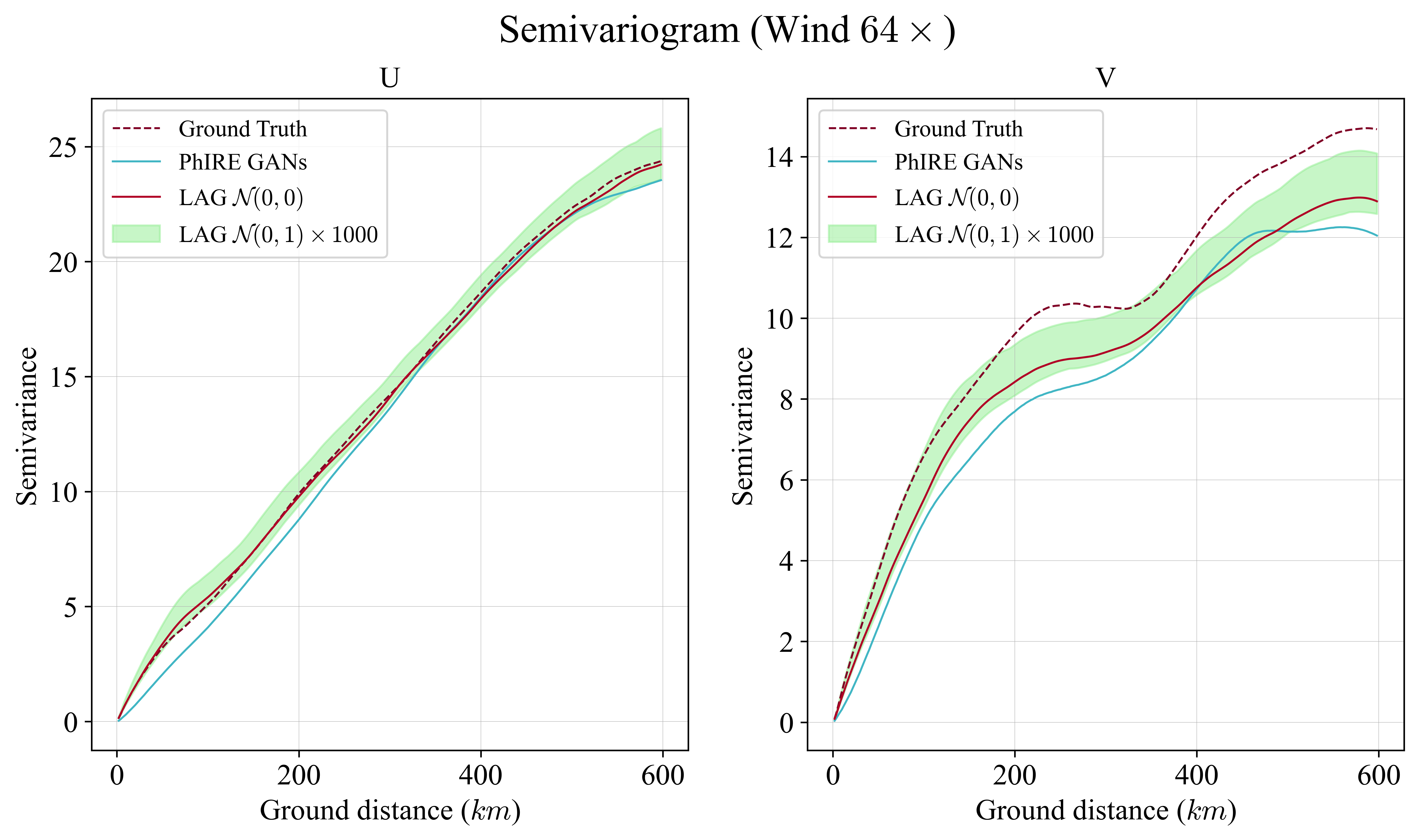}%
    \label{fig:semivariogram_Wind64X}}
  \subfloat[]{\includegraphics[width=.5\columnwidth]{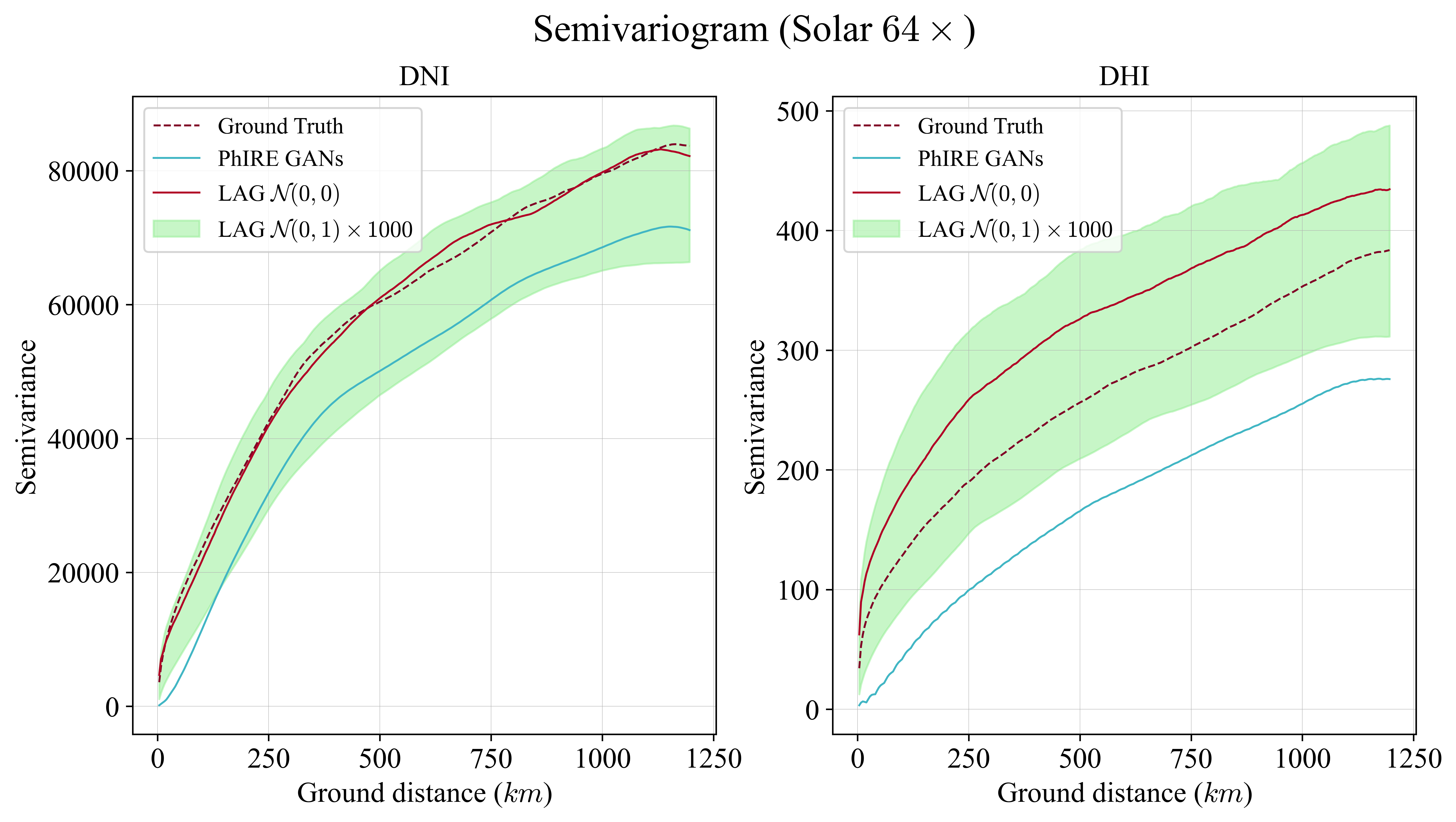}%
    \label{fig:semivariogram_Solar64X}}

  \caption{Semivariograms of the $64\times$ SR results generated by the LAG-based method and the PhIRE GAN for a randomly sampled LR image from (a) wind velocity and (b) solar irradiance test datasets. Images with similar spatial patterns should have close lines. Zoom in for better observation.}

  \label{fig:semivariogram_Wind_Solar_64X}
\end{figure}

Fig.~\ref{fig:semivariogram_Wind_Solar_4X} presents the semivariograms of the ground truth image and the corresponding HR images generated by different models at scaling factor $4$ for a randomly sampled LR image from the wind velocity and solar irradiance test datasets. The red solid line labeled as ``LAG $\mathcal{N}(0,0)$'' represents the semivariogram of the generated image at the center $G(\veczero,\vecy)$. The green envelope represents the empirical interval of the semivariograms of $1000$ realizations produced by the LAG-based method with $\vecz\sim\mathcal{N}(\veczero,\vecone)$. As one can see in Fig.~\ref{fig:semivariogram_Wind_Solar_4X}, EDiffSR and two GAN-based methods (LAG and ESRGAN) produced closer semivariograms to that of the ground truth than ATPK, and DIP, indicating that they well reproduced the spatial patterns of the ground truth data at $4\times$ scaling factor. Among the three outperforming methods, one can see the semivariogram of LAG result is overall the closest to that of the ground truth. It is consistent with the previous visual checking results. Fig.~\ref{fig:semivariogram_Wind_Solar_64X} compares the semivariograms of LAG ($64\times$ SR) and PhIRE GAN ($25\times$ SR for solar irradiance and $50\times$ SR for wind velocity datasets) results. The semivariograms of LAG are closer to that of the ground truth. More semivariogram plots at different scaling factors are given in the Supplementary Material.

\subsection{Mass preservation}\label{sec:mass-preserve}

\begin{figure}
\centering

\subfloat[]{\includegraphics[width=.5\columnwidth]{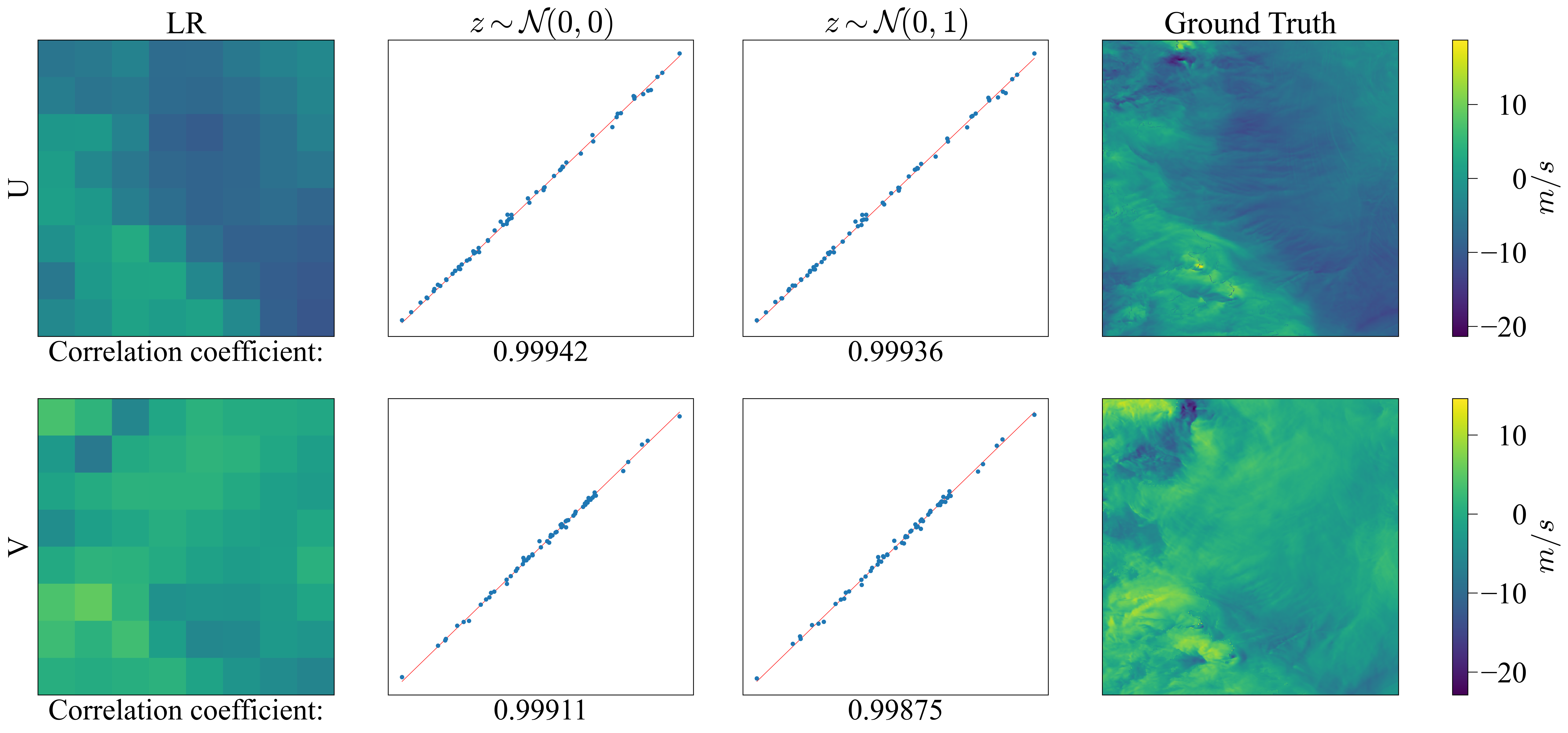}%
\label{fig:mass_preserve_Wind64X_repeat1_ncand2}}
\subfloat[]{\includegraphics[width=.5\columnwidth]{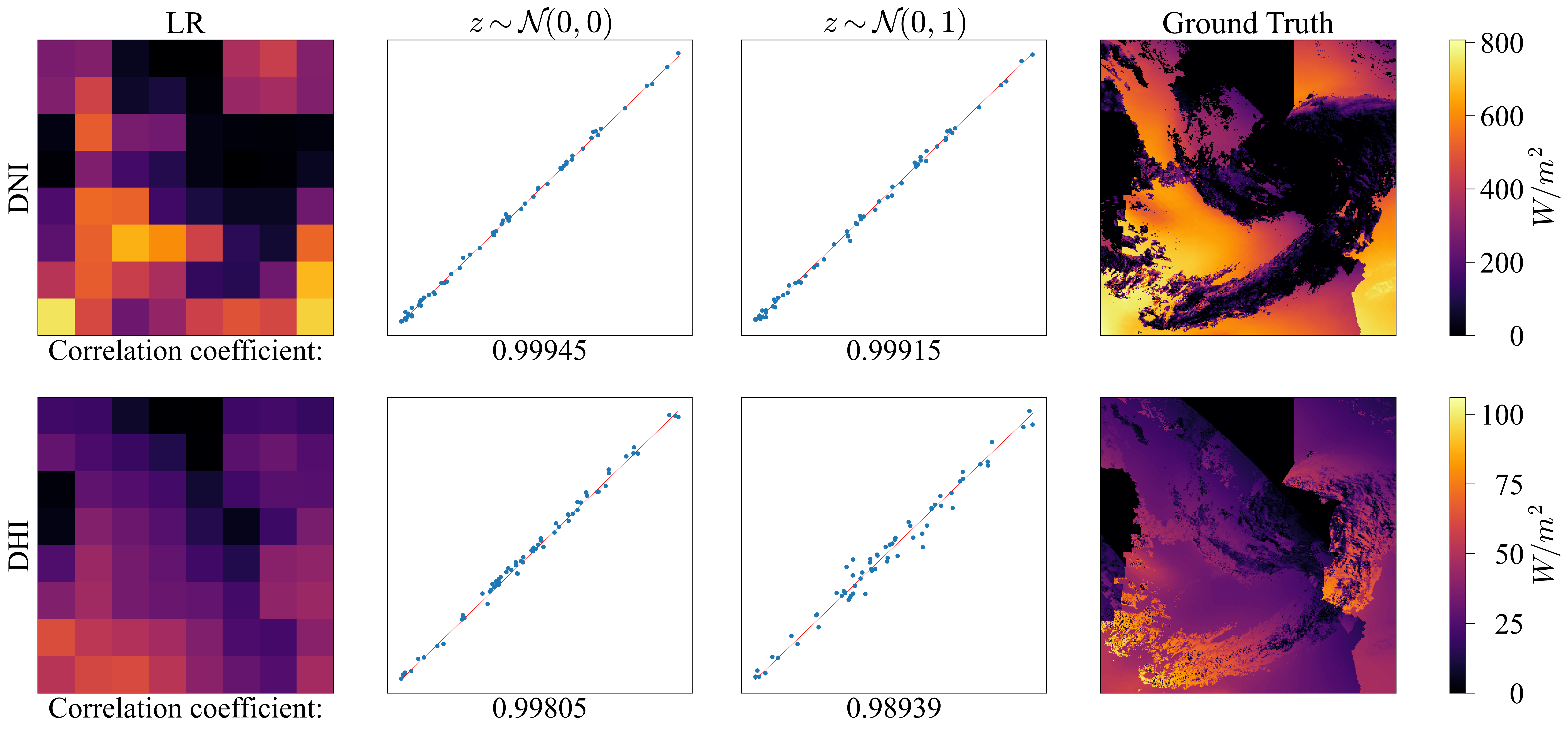}%
\label{fig:mass_preserve_Solar64X_repeat1_ncand2}}

\caption{Consistency check for the $64\times$ SR results generated by the LAG-based method for a randomly sampled LR image from (a) wind velocity and (b) solar irradiance test datasets. A random realization of LAG with $\vecz\sim\mathcal{N}(\veczero,\vecone)$ was used to generate the scatter plots. The horizontal axis represents the pixel values in the input LR image, and the vertical axis represents the average of the associated $64\times64$ pixel values. Scatter points align well with the $45$-degree (red solid) line. The Pearson correlation coefficients (listed at the bottom) are all close to $1$.}

\label{fig:mass_preserve_WindSolar64X_repeat1_nacnd2}
\end{figure}

As mentioned, preserving mass is important for successful scientific downscaling methods. To check the consistency of LAG results, we show scatter plots between the original pixel values and the average of associated pixel values in LAG $64\times$ SR outcomes (image generated with $\vecz = 0$ and a randomly selected image generated with $\vecz\sim\mathcal{N}(\veczero,\vecone)$) (see Fig.~\ref{fig:mass_preserve_WindSolar64X_repeat1_nacnd2}). In the downscaling process with $64\times$ scaling factor, one pixel in an LR image will be refined into $64 \times 64$ pixels. The horizontal axis of each plot represents the pixel values of the input LR image and the vertical axis represents the average values of the corresponding $64\times64$ pixels in the resultant HR images. We can see that the scatter points align well with the $45^\circ$ straight line, suggesting a high degree of consistency between the values of LR pixels and the average of HR pixels. It can also be confirmed by the Pearson correlation coefficients shown at the bottom of each plot, all of which are close to $1$.

\subsection{Uncertainty characterization}\label{sec:hypothesis}

In traditional statistics, the uncertainty of high-dimensional distributions is often examined through Monte Carlo stochastic simulations. This approach involves the generation of numerous random samples from the distribution to approximate and analyze the statistical properties of interest, allowing for the assessment of uncertainty and robustness of complex distributions in an empirical fashion. Similarly, we can leverage the LAG's ability to produce repeated plausible HR images for a downscaling task to empirically characterize the uncertainty of LAG outcomes.

\begin{figure}
\centering

\subfloat[]{\includegraphics[width=.5\columnwidth]{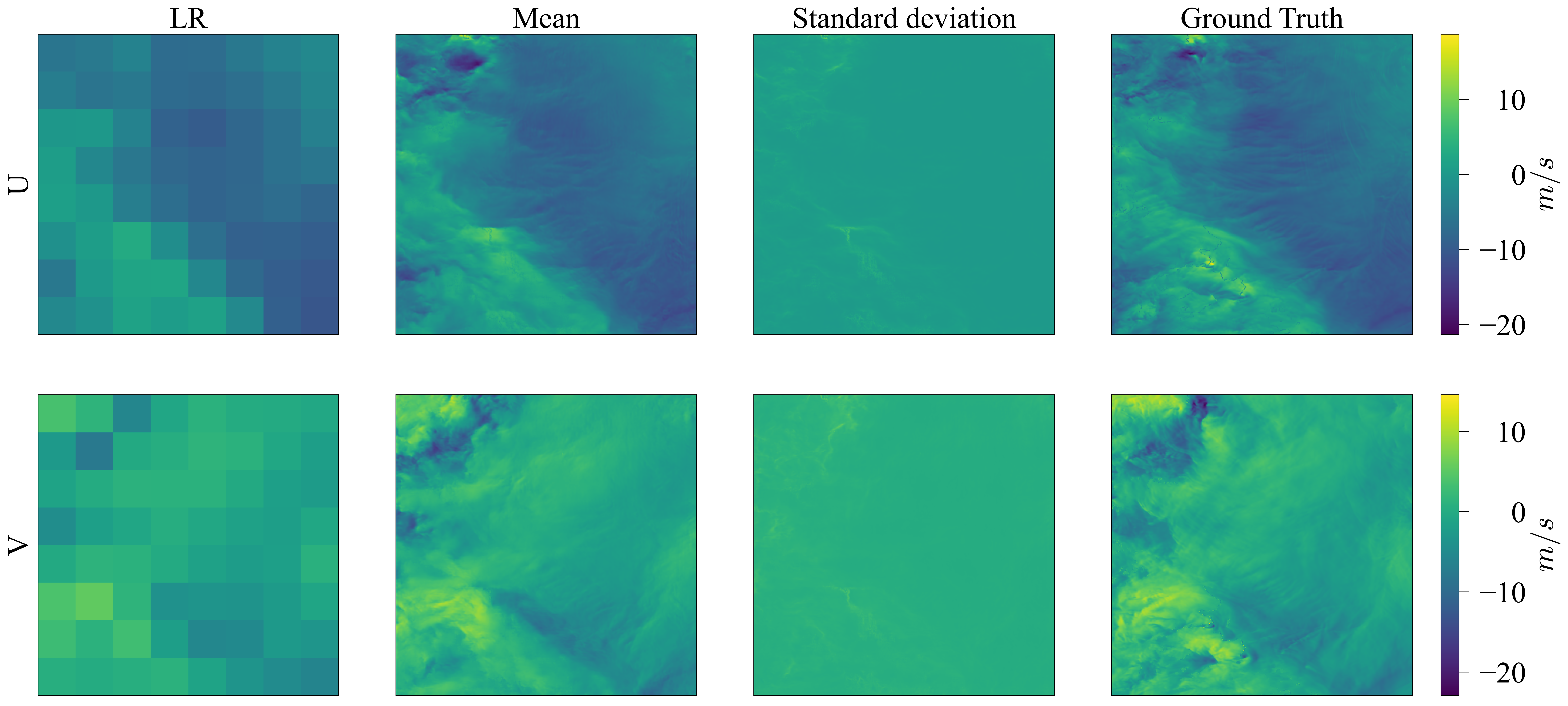}%
\label{fig:testSamples_Wind64X_repeat1000_mean_std}}
\subfloat[]{\includegraphics[width=.5\columnwidth]{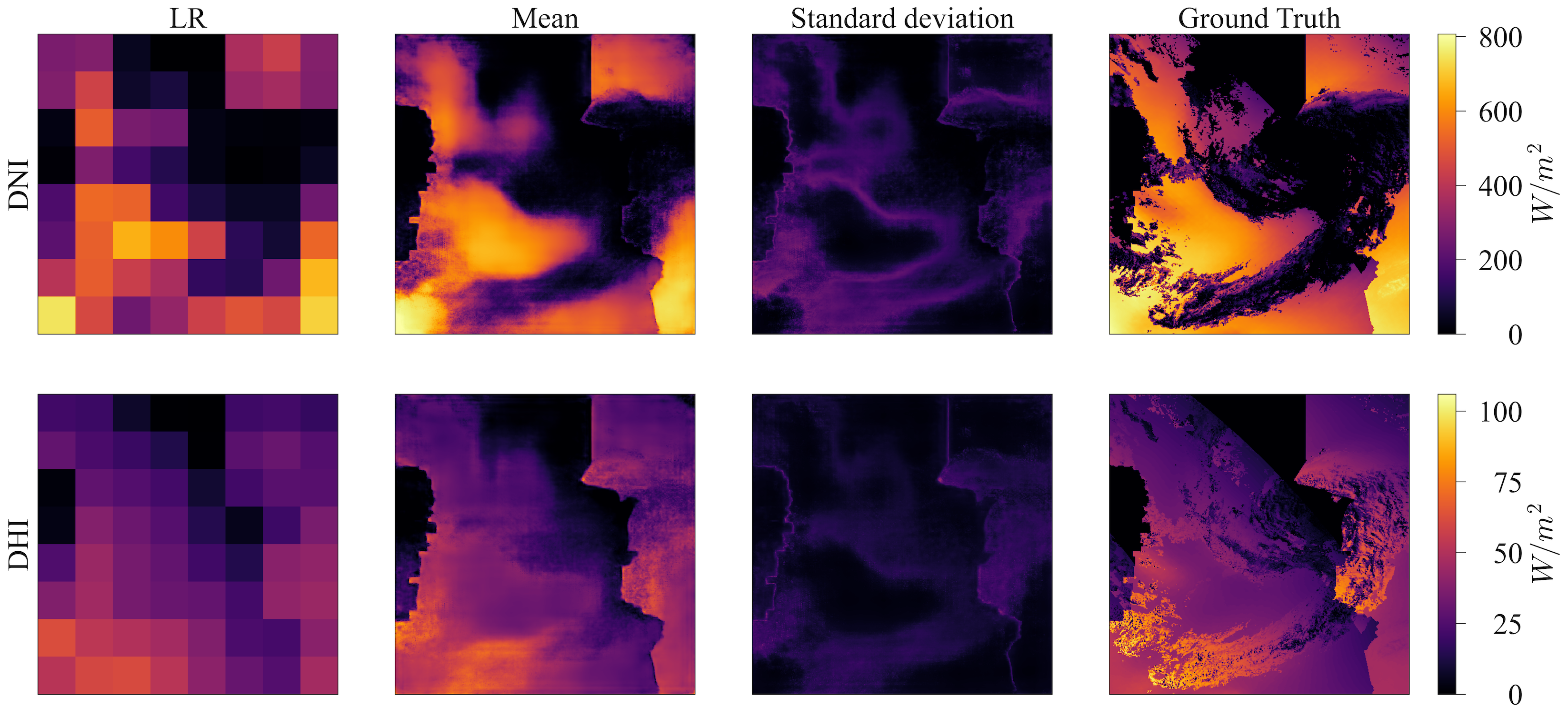}%
\label{fig:testSamples_Solar64X_repeat1000_mean_std}}

\caption{Empirical mean (second column) and standard deviation (third column) images based on the $1000$ random realizations generated by LAG with $\vecz\sim\mathcal{N}(\veczero,\vecone)$ for a randomly sampled LR image from (a) wind velocity and (b) solar irradiance test datasets. For each panel, the leftmost column shows the input LR images at two channels (with image size $8\times8$), the rightmost column shows the corresponding HR ground truth images (with image size $512\times512$). Zoom in for better observation.}
\label{fig:testsamples_windsolar64x_repeat1000_mean_std}
\end{figure}

Fig.~\ref{fig:testsamples_windsolar64x_repeat1000_mean_std} shows the maps of the empirical mean (second column) and standard deviation (third column) for the $64\times$ SR results of randomly sampled LR images (first column) in (a) wind velocity and (b) solar irradiance test datasets. The statistics were derived based on $1000$ realizations of the trained LAG model with $\vecz\sim\mathcal{N}(\veczero,\vecone)$. One can see that estimations with higher standard deviation tend to happen in the edge areas with rapid pixel value changes. The standard deviations for solar irradiance data tend to be higher than that for wind velocity. This can be due to the more complex and heterogeneous spatial patterns in the solar irradiance data. This can also be verified by the semivariogram plots in Figs.~\ref{fig:semivariogram_Wind_Solar_4X} and \ref{fig:semivariogram_Wind_Solar_64X}, where the interval envelopes for solar irradiance data are much wider than those of wind velocity, indicating higher degree of uncertainty in the trained models for solar irradiance data.

We can leverage the simulation-based approach for further statistical inferences. For instance, one can quantify the variation among the random realizations by assessing their divergence from the central tendency of the latent distribution $G(\veczero|\vecy)$. This divergence can be measured using metrics such as residuals, relative MSE, or SWDs. The distribution of the divergence can not only provide further empirical evidence about the model biases and uncertainty, but also facilitate statistical inferences. For example, it allows one to perform hypothesis testing to check if a given HR image could plausibly be a downscaled outcome of an LR image given the information learned in LAG.

\begin{figure}[hbt!]
\centering
\begin{tabular}{lr}
\begin{tabular}{l}
\subfloat[]{\includegraphics[width=.45\columnwidth]{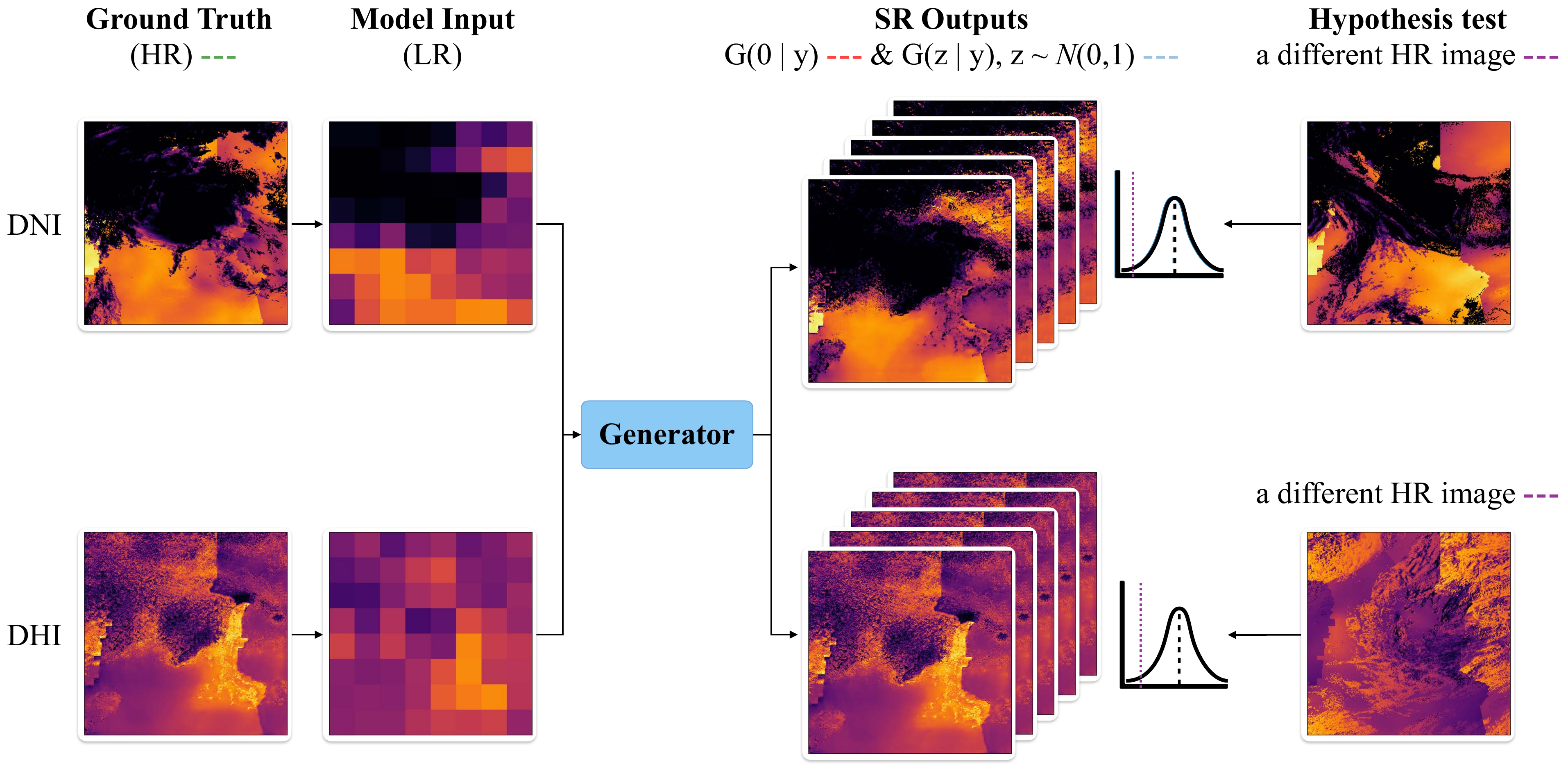}
\label{fig:LAG_HypothesisTest}}
\end{tabular}
&
\begin{tabular}{r}
\subfloat[]{\includegraphics[width=.4\columnwidth]{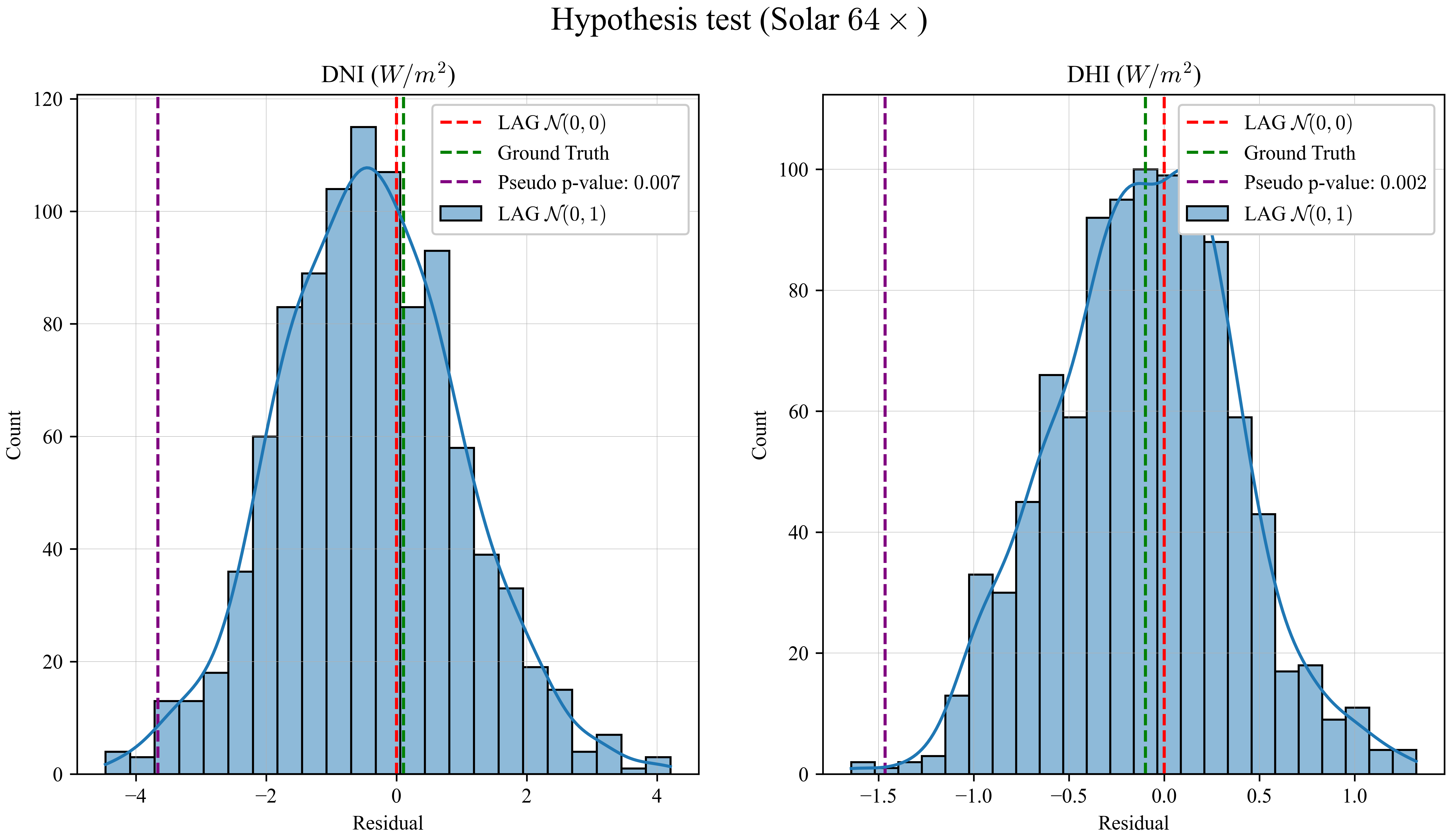}}
\label{fig:LAG_HypothesisTest_residual} \\
\subfloat[]{\includegraphics[width=.4\columnwidth]{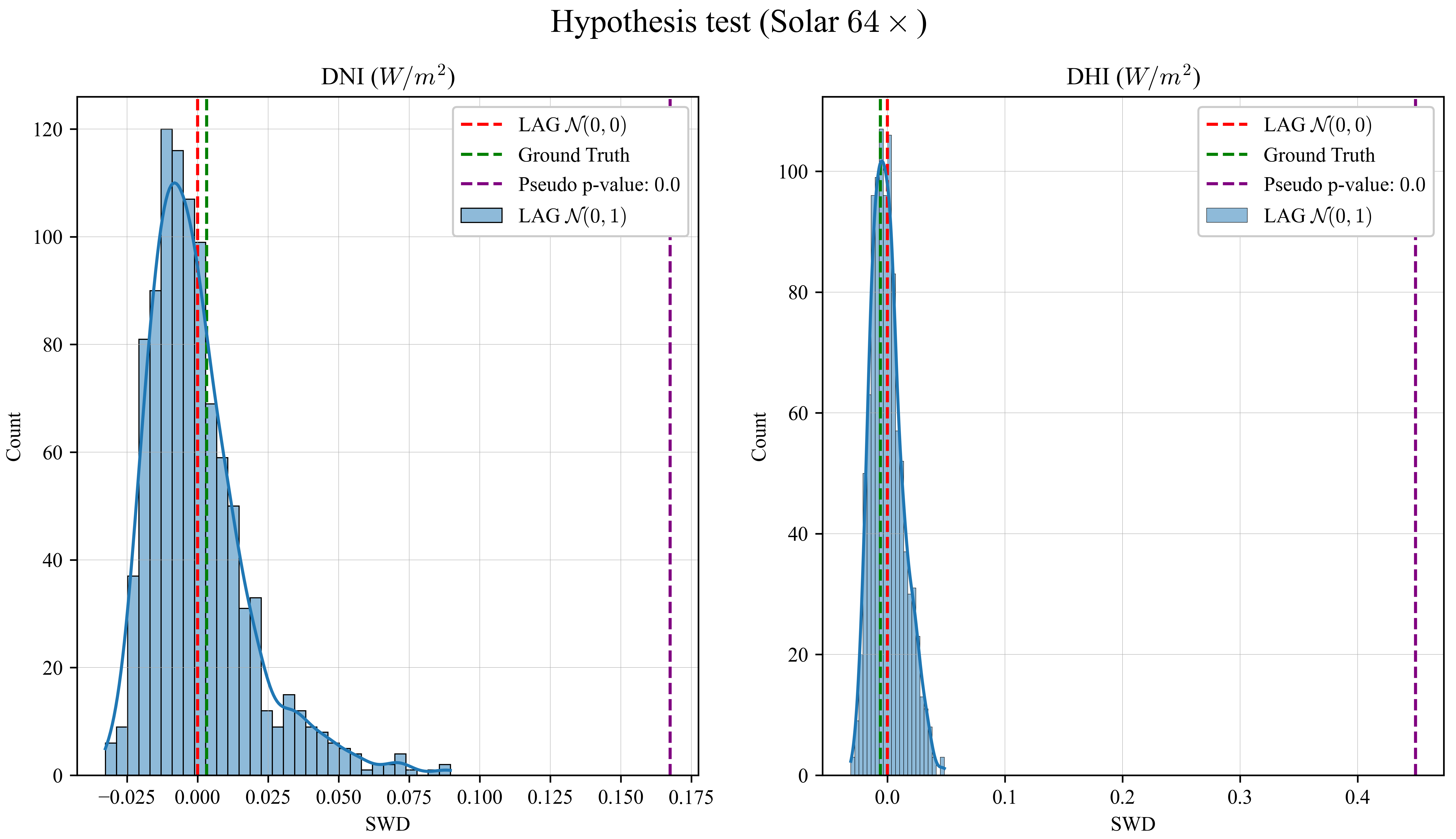}}
\label{fig:LAG_HypothesisTest_swd}
\end{tabular}
\end{tabular}
\caption{An illustration of the LAG based hypothesis testing process using the solar irradiance data (with two channels \textit{DNI} and \textit{DHI}) with $64\times$ scaling factor: (a) shows the diagram of the hypothesis testing process; (b) and (c) are the empirical distributions of the discrepancy of the $1000$ random realizations from the distribution center (red dashed line) measured in residuals and SWD, respectively. The purple dashed line indicates the measure of the HR image to be tested. Zoom in for better observation.}
\label{fig:LAG_HypothesisTest_combine}
\end{figure}

Fig.~\ref{fig:LAG_HypothesisTest_combine} illustrates the process of hypothesis testing with the solar irradiance test dataset. Fig.~\ref{fig:LAG_HypothesisTest_combine}a shows the input LR (with two channels \textit{DNI} and \textit{DHI}), related ground truth images and several random realizations from the trained LAG model. Figs.~\ref{fig:LAG_HypothesisTest_combine}b and \ref{fig:LAG_HypothesisTest_combine}c show the empirical distribution of the discrepancy between each of $1,000$ realizations from the latent distribution center $G(\veczero|\vecy)$ (red dashed line). Given an example HR image (the rightmost in Fig.~\ref{fig:LAG_HypothesisTest_combine}a), we can calculate its discrepancy to the center in terms of residuals or SWD, based on which a pseudo p-value can be derived to statistically test if the given image is a plausible outcome of the LR image. The pseudo p-value is small enough to significantly reject the hypothesis that the given HR image is a plausible downscaled outcome of the LR image. While this example used residuals and SWD to illustrate the concept, it is important to note that other measures of image discrepancy or similarity measures can also be applied to evaluate the relationship between HR and LR images.

\section{Conclusion and Discussions}\label{sec:conclusion}

In this study, we described a LAG-based method for extreme stochastic downscaling for gridded scientific datasets. As a variant of conditional GAN, the LAG method addressed the specific requirements in scientific downscaling, including uncertainty characterization and downscaling with large scaling factors. Compared to existing GAN methods, LAG can explicitly account for the stochasticity in one-to-many mappings inherent to the downscaling process. LAG was implemented with a progressing GAN framework to avoid mode collapse when dealing with large scaling factors. For a given LR image, LAG allows for generating a multitude of plausible downscaling outcomes instead of producing deterministic results. We showcased the advantages of the LAG method with a case study of downscaling gridded climate datasets and highlighted its performances with a comprehensive compression with commonly used downscaling methods developed from different perspectives, including ATPK, DIP, ESRGAN, PhIRE GAN, and EDiffSR. Furthermore, we checked the mass consistency of the LAG output and examined the uncertainty spaces through a simulation-based Monte Carlo approach.

Future improvements of the method are envisioned in two directions. First, we plan to extend the developed downscaling method to spatiotemporal settings, which allows for the generation of time series of gridded datasets with very spatial and temporal resolution. This requires the method to effectively consider the complex spatial patterns and temporal dynamics in the measurements. Secondly, the stochastic nature of the LAG allows for the empirical exploration of the uncertainty space. While this empirical approach has proven effective, adequately incorporating domain knowledge into the modeling process remains challenging. We plan to address this by integrating the LAG with a Bayesian framework \citep{wilson2020case}, which is known for its versatility in incorporating prior knowledge in uncertainty modeling.

\section*{Acknowledgments}
The authors gratefully acknowledge the funding support provided by National Science Foundation (BCS: \#2026331).
This work utilized the Alpine high performance computing resource at the University of Colorado Boulder.
Alpine is jointly funded by the University of Colorado Boulder, the University of Colorado Anschutz,
Colorado State University, and the National Science Foundation (BCS: \#2201538).

\appendix
\section{Supplementary material}

This supplementary material contains additional figures that complement the main article.

\subsection{LAG outputs for different scaling factors}

\begin{figure}
\centering
\includegraphics[width=\columnwidth]{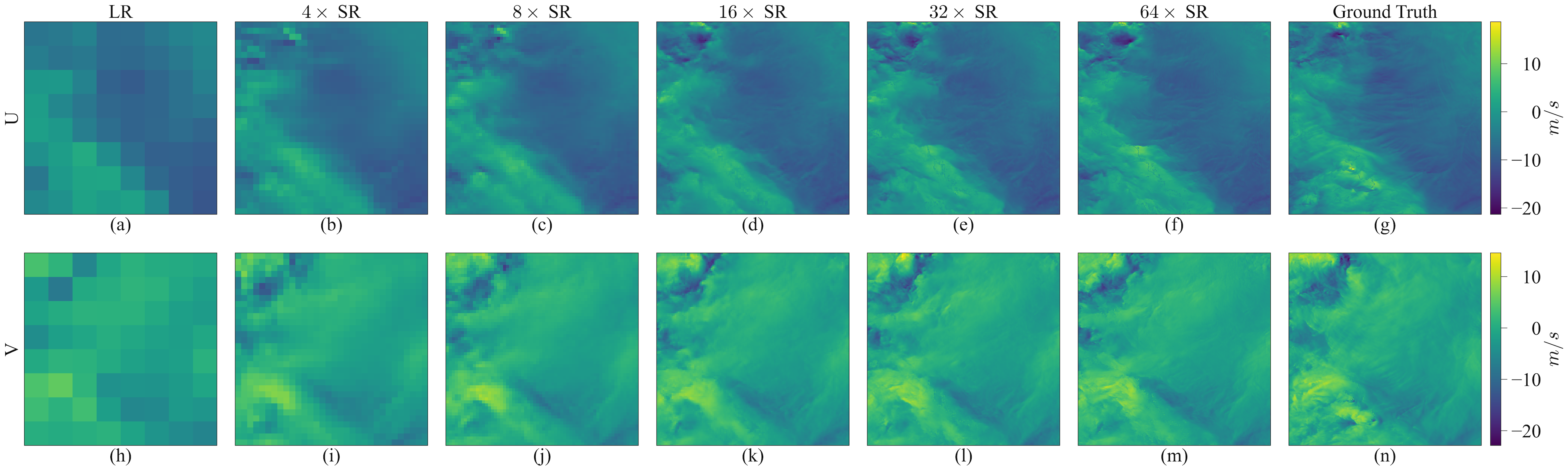}
\caption{SR results generated by the LAG-based downscaling framework for two channels (\textit{U} and \textit{V}) of a randomly sampled LR image from the wind velocity test dataset.}
\label{fig:testSamples_Wind_multipleScale}
\end{figure}

\begin{figure}
\centering
\includegraphics[width=\columnwidth]{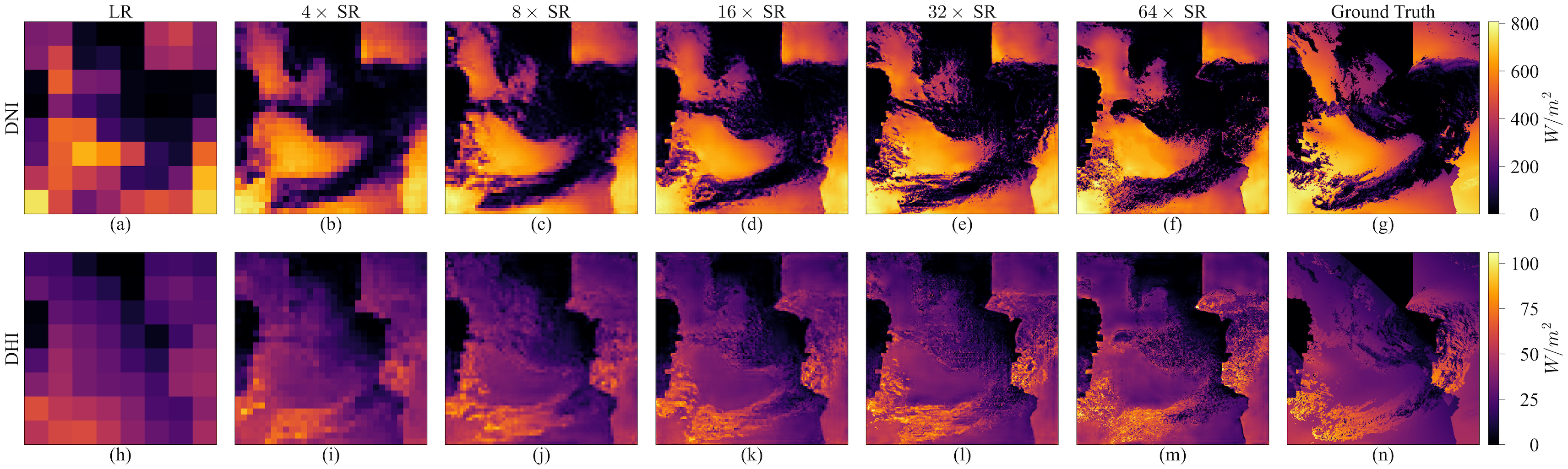}
\caption{SR results generated by the LAG-based downscaling framework for two channels (\textit{DNI} and \textit{DHI}) of a randomly sampled LR image from the solar irradiance test dataset.}
\label{fig:testSamples_Solar_multipleScale}
\end{figure}

Figs. \ref{fig:testSamples_Wind_multipleScale} and \ref{fig:testSamples_Solar_multipleScale} present SR results generated by the LAG-based downscaling framework, $G(\veczero|\vecy)$, for a randomly sampled LR image from the wind velocity and solar irradiance test datasets at different SR scales ($4\times$, $8\times$, $16\times$, $32\times$, and $64\times$). The leftmost column [(a) and (h)] shows the input LR images at two channels (with image size $8 \times 8$), the rightmost column [(g) and (n)] shows the corresponding HR ground truth images (with image size $512 \times 512$). The columns in between shows the SR results at different scaling factors. [(b) and (i)]: $4\times$ SR results with image size $32\times32$; [(c) and (j)]: $8\times$ SR results with image size $64\times64$; [(d) and (k)]: $16\times$ SR results with image size $128\times128$; [(e) and (l)]: $32\times$ SR results with image size $256\times256$; [(f) and (m)]: $64\times$ SR results with image size $512\times512$.

\begin{figure}
\centering
\subfloat[]{\includegraphics[width=.5\columnwidth]{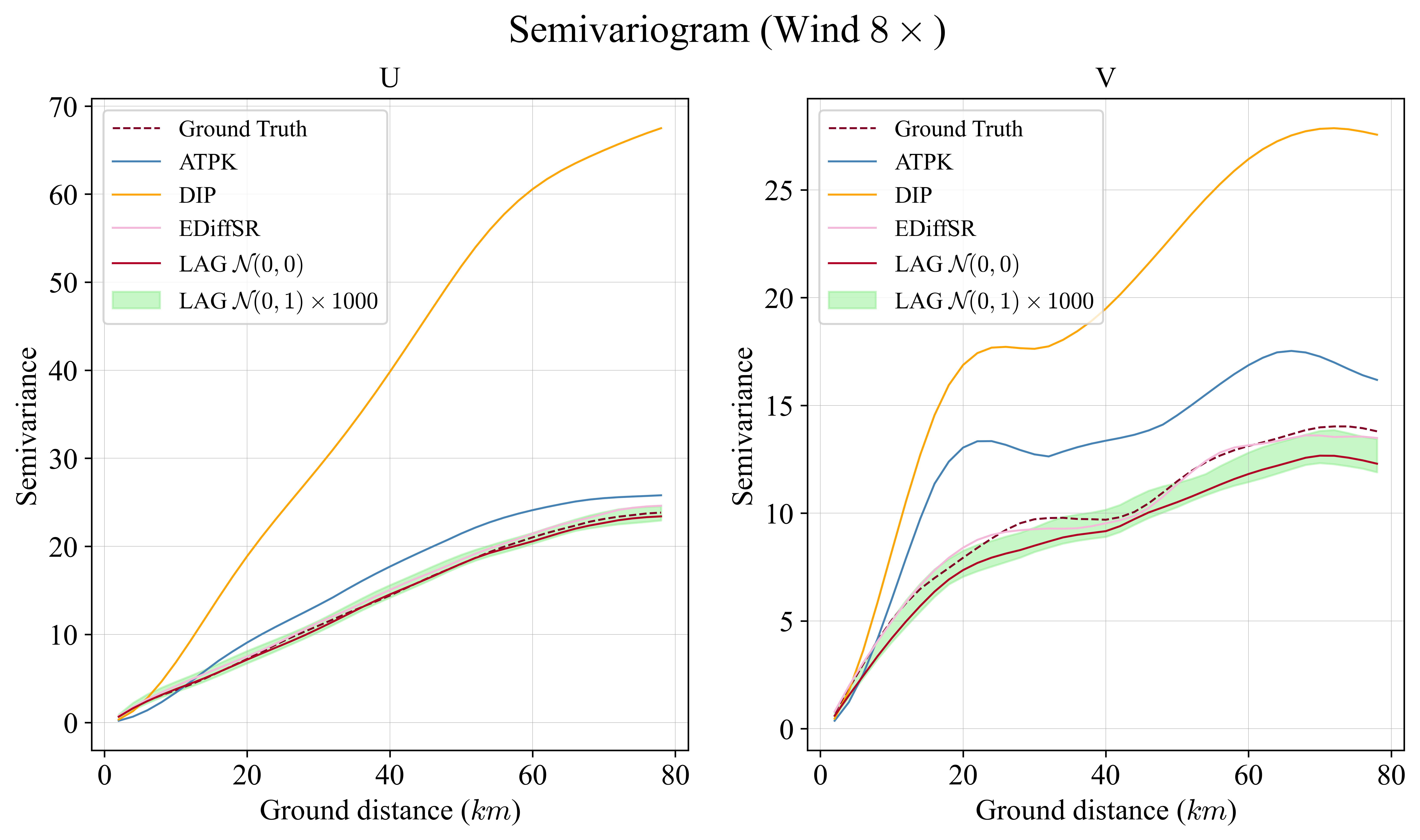}%
\label{fig:semivariogram_Wind8X}}
\subfloat[]{\includegraphics[width=.5\columnwidth]{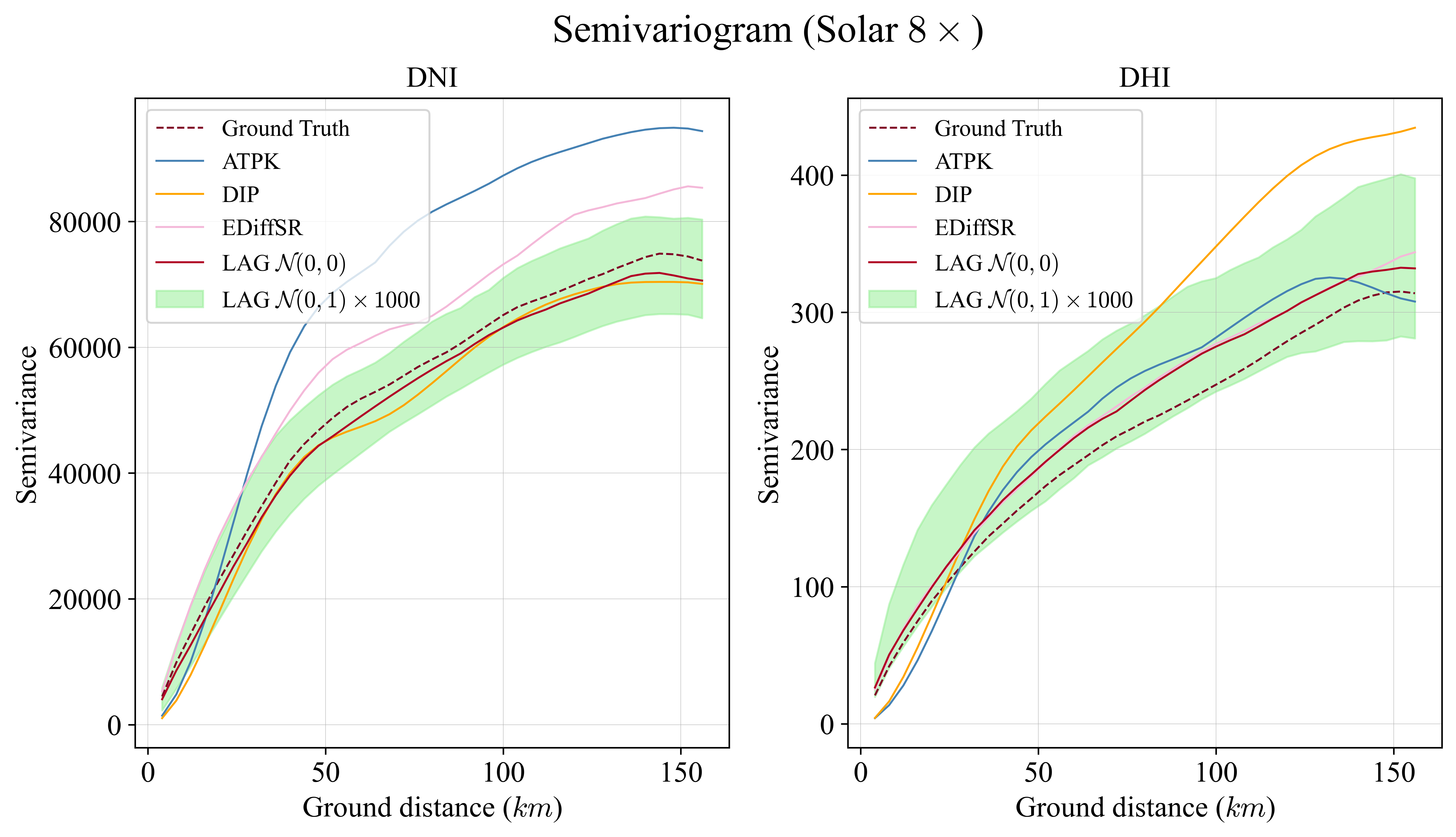}%
\label{fig:semivariogram_Solar8X}}
\caption{Semivariograms of the $8\times$ SR results generated by different models for a randomly sampled LR image from (a) wind velocity and (b) solar irradiance test datasets.}
\label{fig:semivariogram_Wind_Solar_8X}
\end{figure}

\begin{figure}
\centering
\subfloat[]{\includegraphics[width=.5\columnwidth]{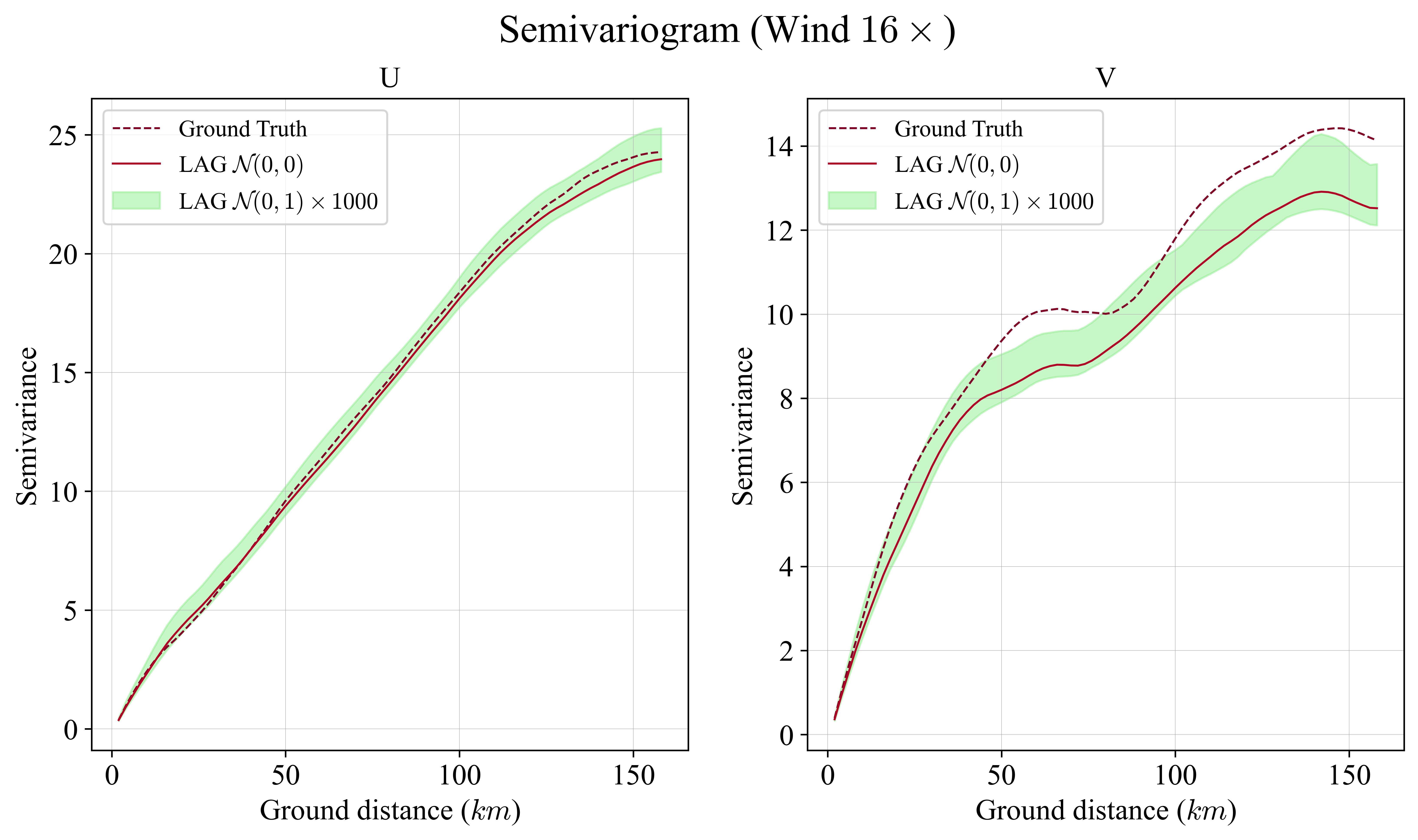}%
\label{fig:semivariogram_Wind16X}}
\subfloat[]{\includegraphics[width=.5\columnwidth]{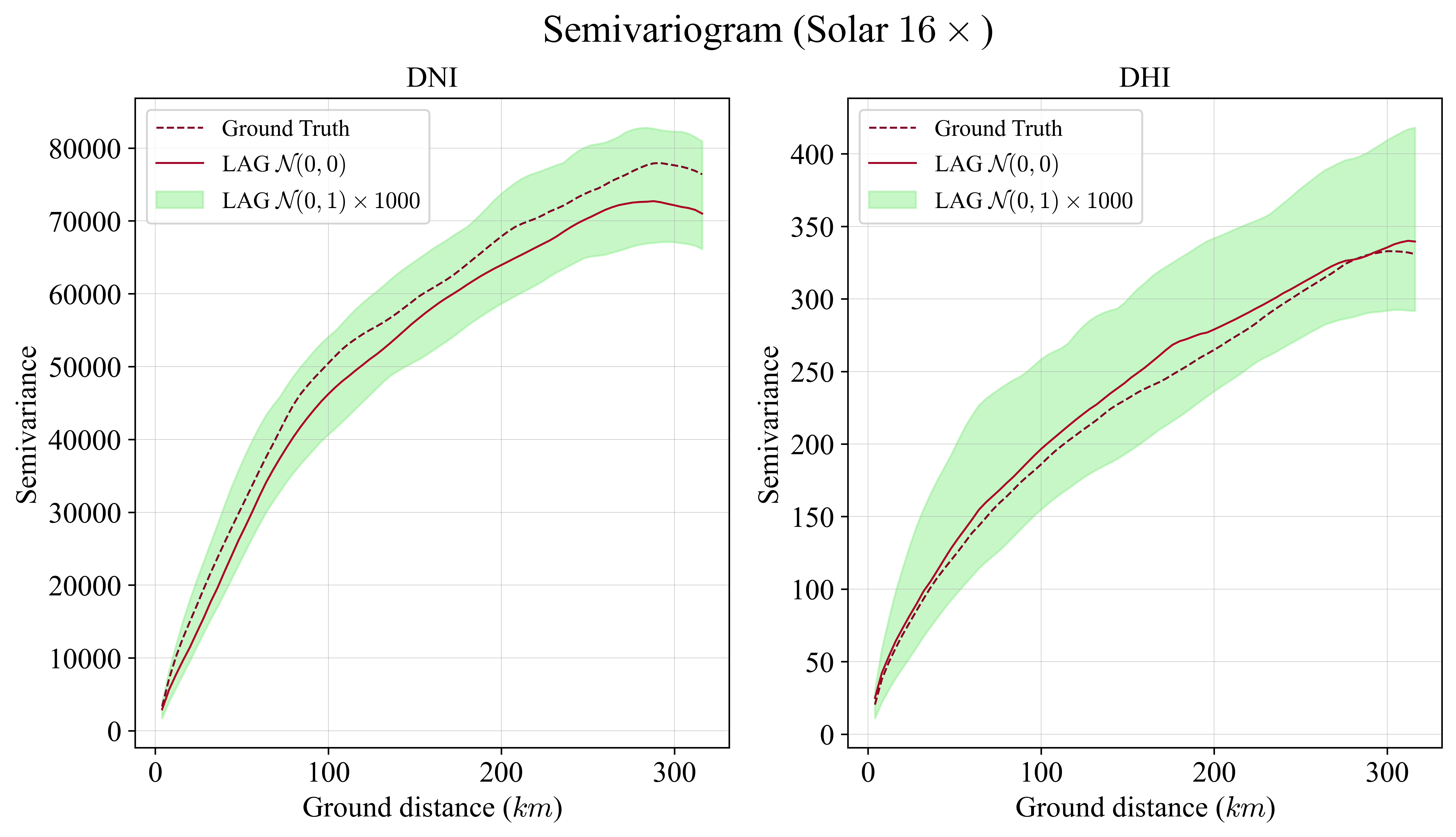}%
\label{fig:semivariogram_Solar16X}}
\caption{Semivariograms of the $16\times$ SR results generated by the LAG-based downscaling framework for a randomly sampled LR image from (a) wind velocity and (b) solar irradiance test datasets.}
\label{fig:semivariogram_Wind_Solar_16X}
\end{figure}

\begin{figure}
\centering
\subfloat[]{\includegraphics[width=.5\columnwidth]{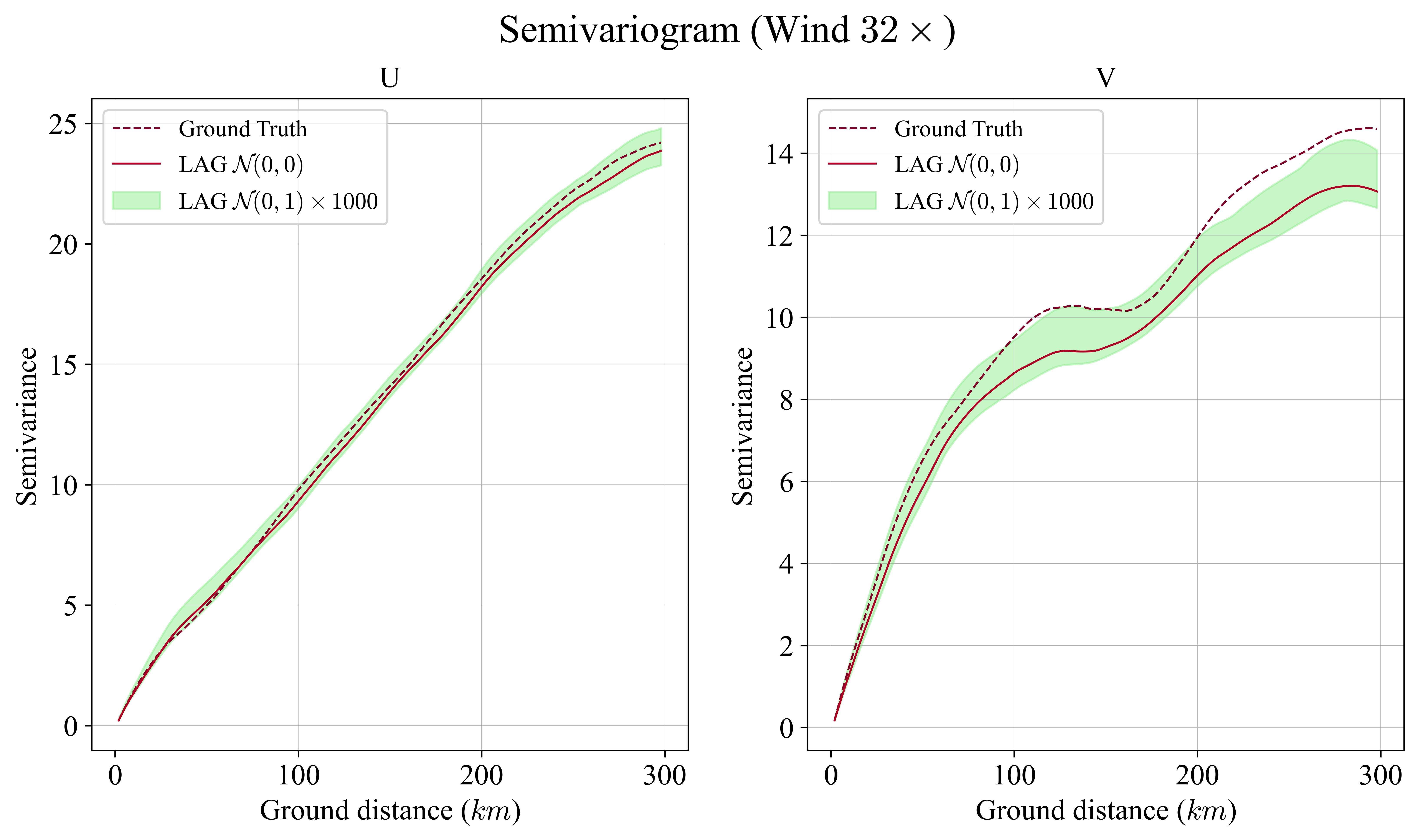}%
\label{fig:semivariogram_Wind32X}}
\subfloat[]{\includegraphics[width=.5\columnwidth]{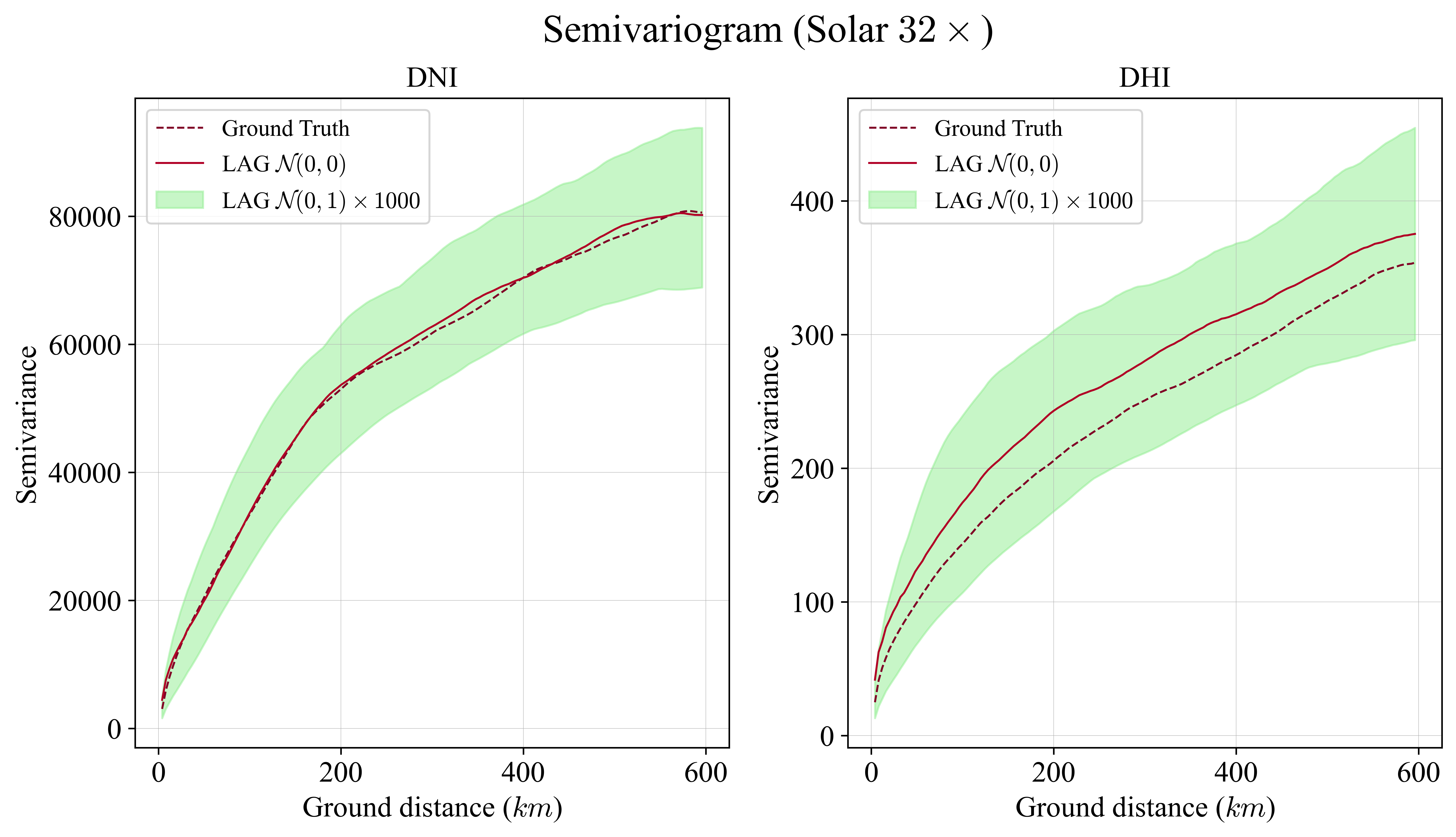}%
\label{fig:semivariogram_Solar32X}}
\caption{Semivariograms of the $32\times$ SR results generated by the LAG-based downscaling framework for a randomly sampled LR image from (a) wind velocity and (b) solar irradiance test datasets.}
\label{fig:semivariogram_Wind_Solar_32X}
\end{figure}

Figs. \ref{fig:semivariogram_Wind_Solar_8X}, \ref{fig:semivariogram_Wind_Solar_16X}, and \ref{fig:semivariogram_Wind_Solar_32X} present the corresponding semivariograms of the SR results ($8\times$, $16\times$, and $32\times$) generated by LAG and comparison models.

\subsection{Error distribution for LAG}
\begin{figure}
\centering
\subfloat[]{\includegraphics[width=.49\columnwidth]{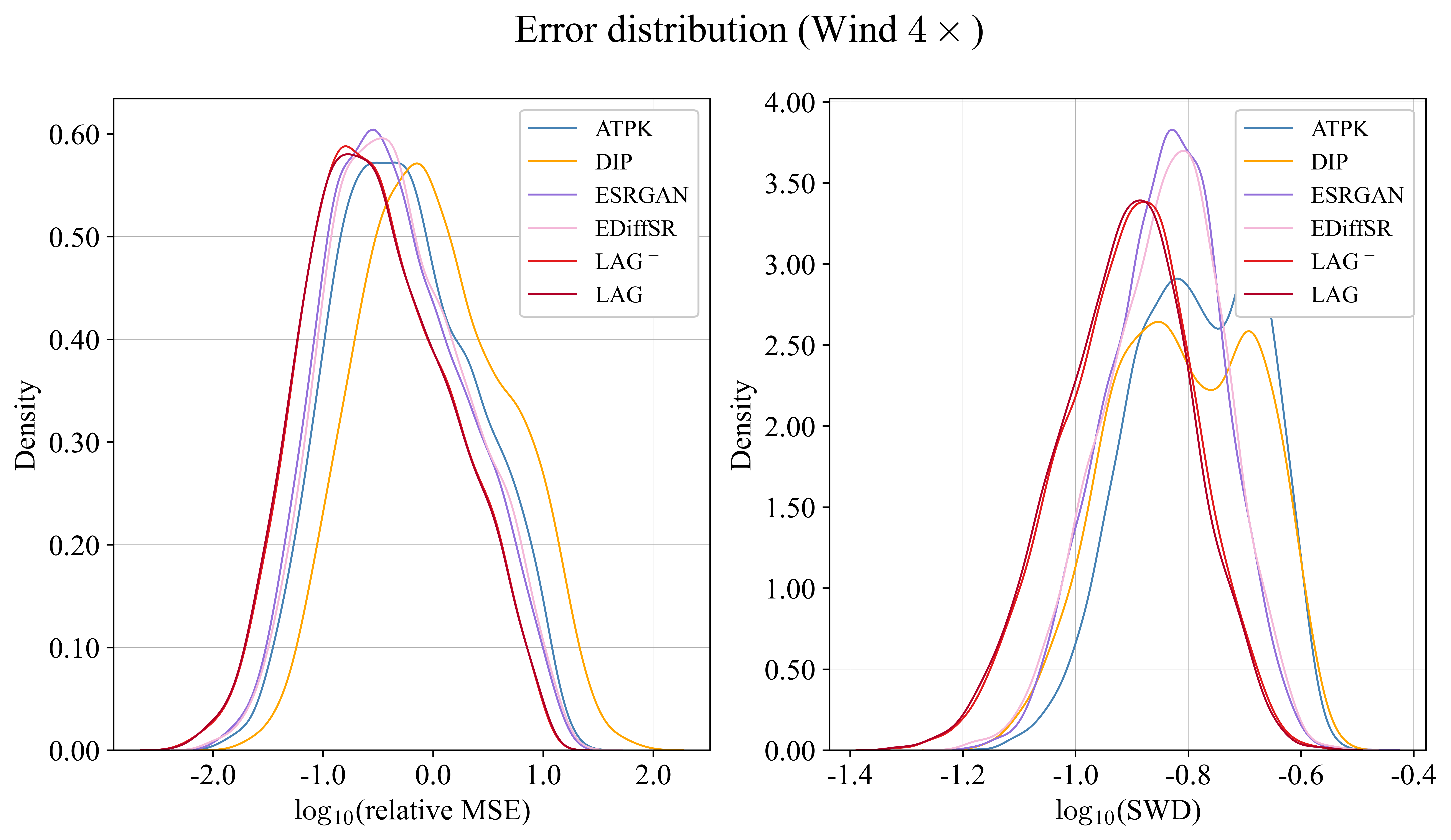}
\label{fig:error_hist_Wind_4X}}
\subfloat[]{\includegraphics[width=.49\columnwidth]{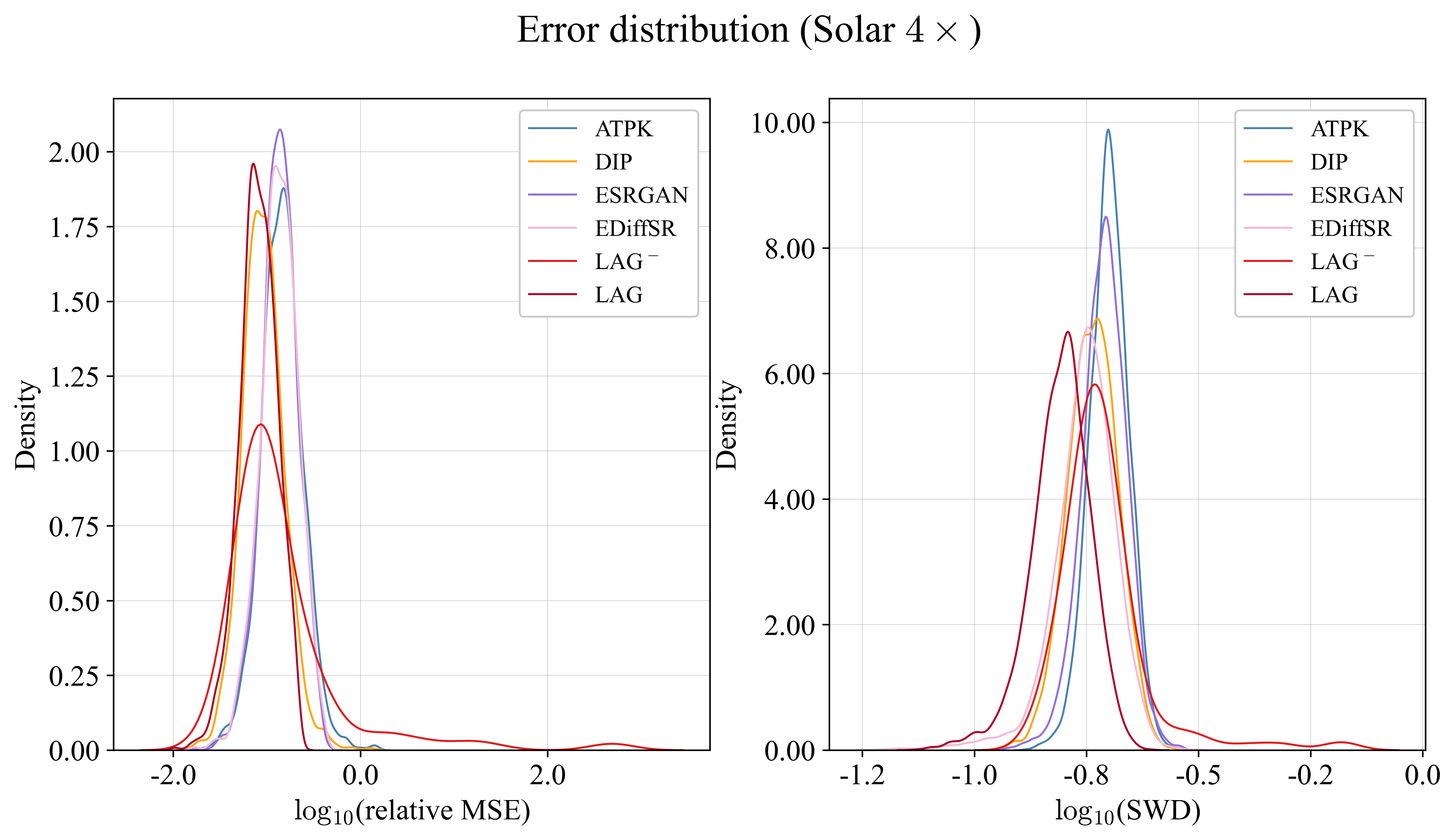}
\label{fig:error_hist_Solar_4X}}
\caption{Empirical density distribution of the relative MSE and SWD in logarithmic scale across (a) wind velocity and (b) solar irradiance test datasets of different models at $\textbf{4}\times$ SR scale.}
\label{fig:error_hist_Wind_Solar_4X}
\end{figure}

\begin{figure}
\centering
\subfloat[]{\includegraphics[width=.49\columnwidth]{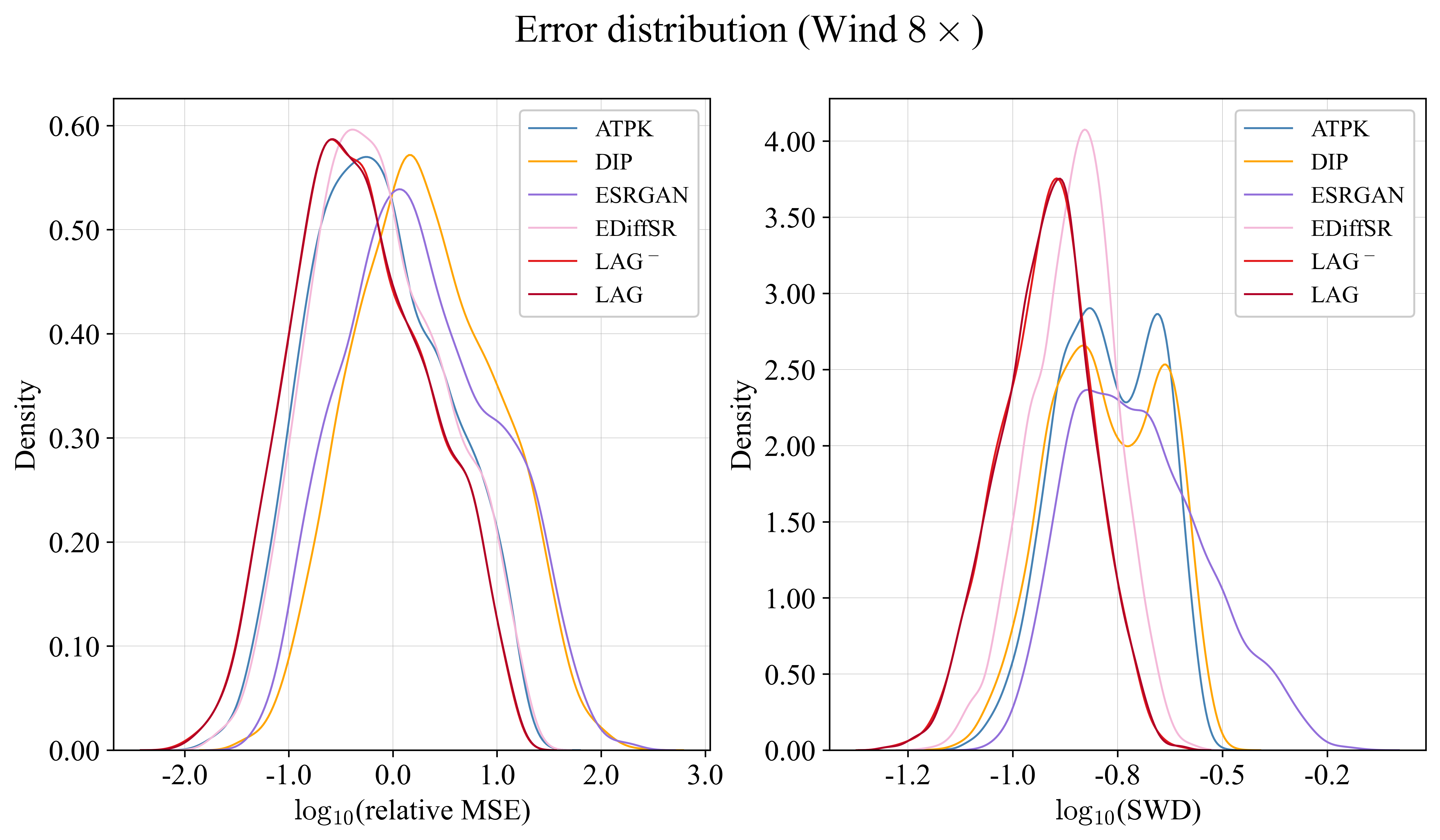}
\label{fig:error_hist_Wind_8X}}
\subfloat[]{\includegraphics[width=.49\columnwidth]{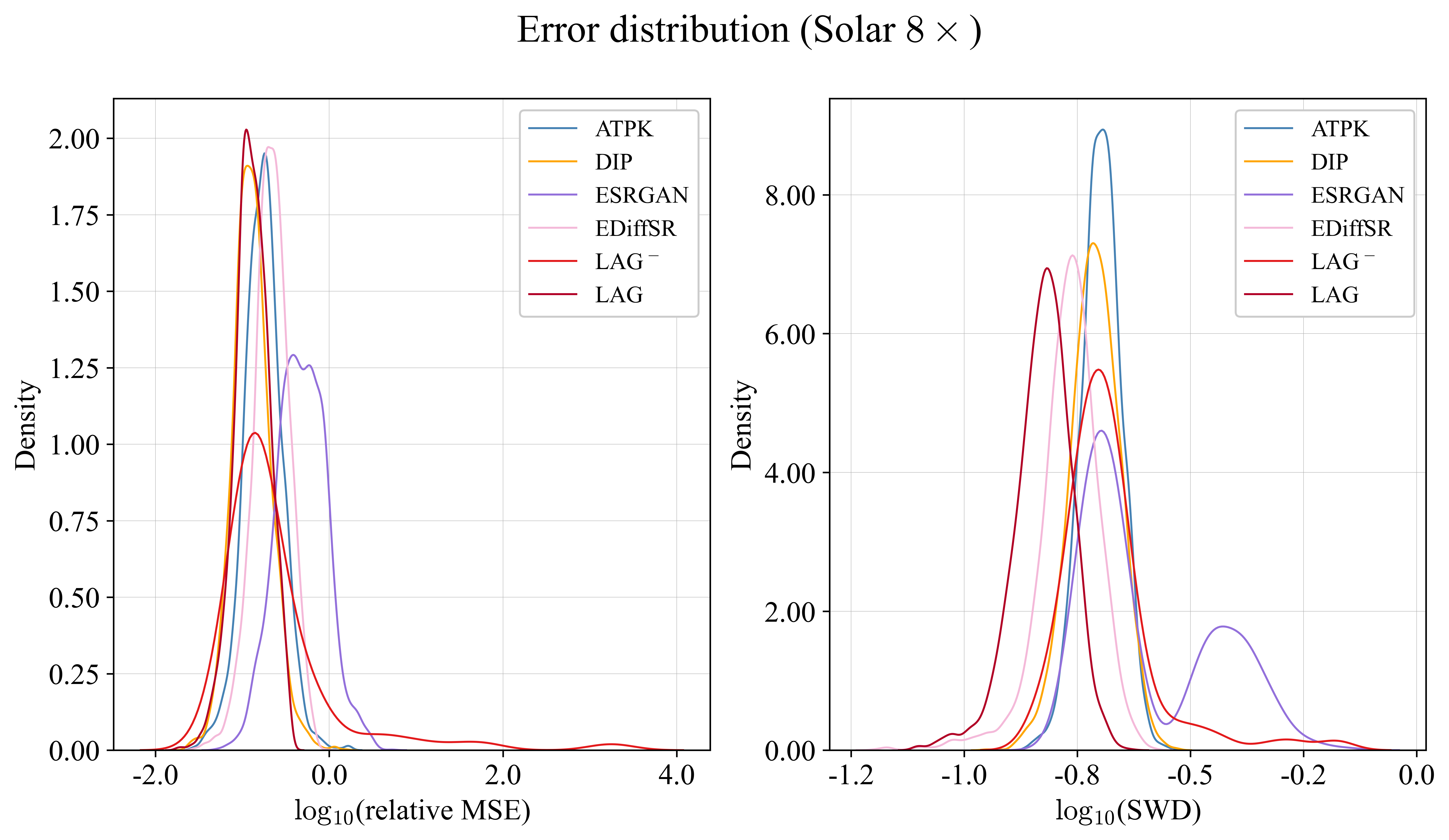}
\label{fig:error_hist_Solar_8X}}
\caption{Empirical density distribution of the relative MSE and SWD in logarithmic scale across (a) wind velocity and (b) solar irradiance test datasets of different models at $\textbf{8}\times$ SR scale.}
\label{fig:error_hist_Wind_Solar_8X}
\end{figure}

Fig.~\ref{fig:error_hist_Wind_Solar_4X} shows the empirical density distribution of the relative MSE and SWD on a logarithmic scale across the wind velocity and solar irradiance test datasets of different models at $4\times$ SR scale. LAG-based models outperform other models. As the scaling factor increases, the advantages of our model become even clearer; see Fig.~\ref{fig:error_hist_Wind_Solar_8X} for the comparison of $8\times$ results. Zoom in for better observation.

\newpage
\bibliographystyle{elsarticle-harv}
\bibliography{LAG_JAG}

\end{document}